\documentclass{article}

\PassOptionsToPackage{numbers, compress}{natbib}
 \usepackage[preprint]{neurips_2026}

\usepackage[utf8]{inputenc} 
\usepackage[T1]{fontenc}    
\usepackage{hyperref}       
\usepackage{url}            
\usepackage{booktabs}       
\usepackage{amsfonts}       
\usepackage{nicefrac}       
\usepackage{microtype}      
\usepackage{xcolor}         

\usepackage{graphicx}
\usepackage{wrapfig}
\usepackage{amsmath}
\usepackage{multirow}
\usepackage{pifont}
\usepackage{amssymb}

\newcommand{\paragrapht}[1]{\noindent\textbf{#1}}
\newcommand{\ours}{C4G}

\title{Learning Global Motion with Compact Gaussians for Feed-Forward 4D Reconstruction}

\author{Mungyeom Kim$^{1}$\footnotemark[1] \; Minkyeong Jeon$^{1}$\footnotemark[1] \; Honggyu An$^1$\footnotemark[1]  \; Jaewoo Jung$^1$ \; \\ \textbf{Hyunah Ko}$^{1}$ \; \textbf{Jisang Han}$^1$ \; \textbf{Hyeonseo Yu}$^1$ \;  \textbf{Donghwan Shin}$^1$ \; \textbf{Sunghwan Hong}$^2$ \\
\textbf{Takuya Narihira}$^3$ \; \textbf{Kazumi Fukuda}$^3$ \;
\textbf{Yuki Mitsufuji}$^{3,4}$\footnotemark[2] \;  \textbf{Seungryong Kim}$^{1}$\footnotemark[2] \\[5pt] $^1$ KAIST AI  \,\,$^2$ ETH Z\"urich, ETH AI Center \,\, $^3$ Sony AI \,\, $^4$ Sony Group Corporation\\[5pt]
\tt\ \small \textcolor{blue!60!black}{\href{https://cvlab-kaist.github.io/C4G}{https://cvlab-kaist.github.io/C4G}}
\vspace{-10pt}
}

\begin{document}

\maketitle

\begin{abstract}
Dynamic scene reconstruction from monocular video remains a fundamental challenge in computer vision. Existing feed-forward methods predict 3D Gaussians pixel-wise for each frame, suffering from duplicated Gaussians and view-dependent biases that hinder effective learning of scene motion. We present \ours, a feed-forward 4D reconstruction framework built upon a compact set of timestamp-conditioned learnable Gaussian query tokens. Each token aggregates corresponding features across the full temporal context and decodes a 3D Gaussian whose position is modulated by the target timestamp, enabling globally coherent motion modeling without per-scene optimization. To capture fine-grained details, we further introduce a video diffusion model-based rendering enhancement module. Since our framework effectively aggregates features into Gaussians, we extend this capability to feature lifting, producing a 4D feature field that supports point tracking and dynamic scene understanding. \ours\ achieves strong novel-view synthesis performance using significantly fewer Gaussians and without requiring camera poses, while exhibiting stronger motion modeling and robustness to large temporal gaps.
\end{abstract}
\section{Introduction}
\label{sec:intro}

Dynamic 4D reconstruction from monocular video is a fundamental challenge in computer vision with broad implications for AR/VR simulation~\cite{pan2023aria}, robotics~\cite{nair2022r3m,lee2026tora}, and content creation~\cite{yang2026neoverse}. Recent extensions of 3D Gaussian Splatting (3DGS)~\cite{kerbl20233d} to dynamic scenes have achieved high-fidelity 4D reconstruction through designs such as deformable fields~\cite{yang2024deformable}, or motion modelling~\cite{lei2025mosca, wang2025freetimegs, wang2025shape}.
However, these methods rely on per-scene optimization, requiring lengthy optimization for every new video and fundamentally limiting their scalability to general-purpose video inputs.

Following the recent success of feed-forward 3DGS~\cite{charatan2024pixelsplat, zhang2024gs}, several works have sought to overcome the limitations of per-scene optimization by learning generalizable feed-forward models for dynamic scenes~\cite{xu20254dgt, yang2026neoverse, liang2024feed, lin2025movies, hurufo}. Given monocular video as input, they estimate per-pixel 3D Gaussians in a feed-forward manner, either reconstructing the scene at specific timestamp~\cite{liang2024feed, shen2025seeing} or augmenting each Gaussian with attributes such as velocity or flow to model the full 4D scene~\cite{hurufo, xu20254dgt, yang2026neoverse}.

\begin{figure}[t]
    \centering
    \includegraphics[width=0.95\linewidth]{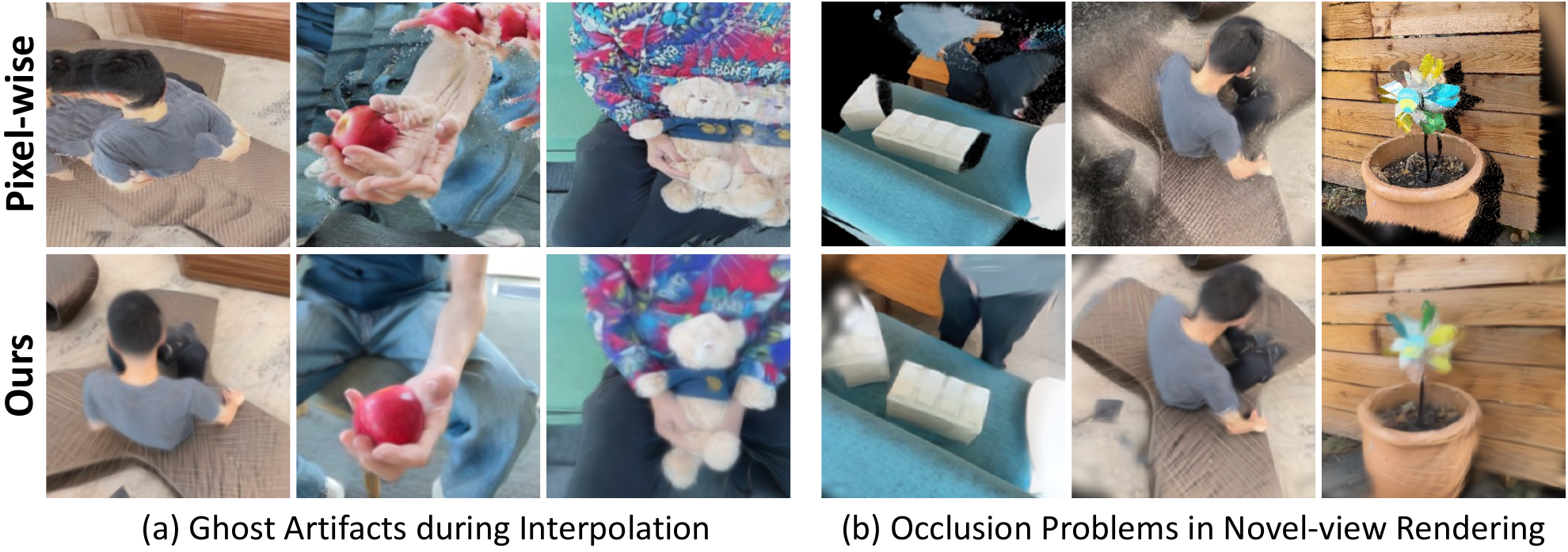}
    \vspace{-5pt}
    \caption{\textbf{Failures of pixel-wise feed-forward 4D reconstruction~\cite{xu20254dgt, lin2025movies, yang2026neoverse}.} (a) Duplicated Gaussians from nearby input views cause ghost artifacts at target timestamps. (b) View-dependent bias prevents leveraging temporally distant views, leaving occluded regions poorly reconstructed.
    }
    \label{fig:failure_modes}
    \vspace{-15pt}
\end{figure}

Despite their effectiveness on 4D reconstruction, we find that these methods exhibit critical failure modes when handling \textit{large temporal motion}, as illustrated in Fig.~\ref{fig:failure_modes}-(a). When rendering at an interpolated timestamp between input frames, per-pixel methods produce spatially duplicated Gaussians---one set from each nearby input view---resulting in ghosting artifacts and degraded rendering quality.
Moreover, even when information from temporally distant views is necessary to fill occluded regions (Fig.~\ref{fig:failure_modes}-(b)), these methods exhibit a strong \emph{view-dependent bias}~\cite{kim20244d}: rendering at a given timestamp relies predominantly on Gaussians from the temporally closest input view, failing to leverage globally available information and leaving occluded regions poorly reconstructed.

We argue that both issues stem from the fundamental design choice shared by all existing feed-forward 4D methods: \textit{per-pixel Gaussian prediction}.
In the static setting, recent analyses~\cite{an2025c3g, miao2025evolsplat, wang2025volsplat} have already shown that pixel-aligned prediction provides excessive representational capacity, causing models to overfit to input-view distributions rather than learn globally consistent geometry from photometric supervision. This problem is further exacerbated in the 4D setting: since consecutive video frames exhibit smooth, gradual motion, a per-pixel model can achieve high rendering scores simply by interpolating Gaussians from nearby input views, without any incentive to learn the true underlying motion or geometry of the scene.

To address this, we propose Compact 4D representation with Gaussians (\textbf{\ours{}}), a feed-forward dynamic scene reconstruction framework that replaces per-pixel prediction with \textbf{learnable query-based prediction}. Following prior work~\cite{yang2026neoverse, ye2024no}, we extract multi-frame features with a geometric foundation model~\cite{wang2025vggt}. We then introduce a compact set of learnable queries that aggregate temporally and geometrically consistent information across the input frames via a transformer decoder, and decode them into 3D Gaussians. For temporal reasoning, we inject per-frame timestamp embeddings into the features and condition the queries on a target timestamp. This allows the decoded Gaussians to represent the scene at any moment, and dynamic scenes are obtained by decoding with varying target timestamps. By keeping the query set compact and aggregating across the entire input sequence, our design forces the Gaussians to capture \textit{global motion} rather than overfitting to individual frames.



While keeping the number of Gaussians compact is essential for the model to learn global motion, it inevitably introduces a trade-off with rendering quality. Although previous approaches~\cite{an2025c3g, jiang2025anysplat} rely on 3DGS optimization~\cite{lee20253d}, it is inapplicable to dynamic scenes due to the lack of multi-view consistency. We therefore propose a rendering enhancement module based on a Video Diffusion Model (VDM)~\cite{jiang2025vace}, which we fine-tune to refine rendered frames conditioned on the input views. This VDM then serves as rendering enhancer that directly improves the visual quality of renderings.



After training, each learnable query fetches multi-frame features from spatially coherent regions around its corresponding Gaussian, retrieving only the information necessary to represent global motion. Building on this property, we propose feed-forward feature lifting network that produces \textbf{4D feature fields}, enabling point tracking and dynamic scene understanding. Whereas prior feed-forward 4DGS methods struggle to lift VFM features due to the input-view bias problem, our feature lifting decoder reuses the same learnable queries and attention patterns in \ours{} decoder to lift VFM features~\cite{simeoni2025dinov3, wang2025vggt} into 3D Gaussians, aggregating only those spatially coherent with each Gaussian. Since each Gaussian and its feature attribute originate from the same token through the same attention pattern, the attribute can be directly attached to its Gaussian, yielding a 4D feature field.

To validate our framework, we conduct extensive novel-view synthesis experiments on multiple dynamic benchmarks~\cite{gao2022monocular, pan2023aria, sturm2012benchmark, yoon2020novel}. Although our compact representation uses \textbf{0.007 $\times$} or fewer Gaussians than per-pixel methods~\cite{xu20254dgt, lin2025movies, yang2026neoverse}, it achieves state-of-the-art or competitive performance across all datasets by correctly modeling scene motion and aggregating information from the full temporal context. We further evaluate Gaussian motion estimation via 2D point tracking to validate our model understand the global motion, where our method substantially outperforms per-pixel Gaussian approaches. Finally, our 4D feature field is assessed on point tracking and dynamic scene understanding, confirming that the model produces meaningful 4D representations.



\section{Related Works}
\label{sec:relwork}

\paragrapht{Per-scene Optimization-based 4D Reconstruction.} 
Given a monocular video, traditional methods reconstruct non-rigid contents through RGBD sensors~\cite{bozic2020deepdeform,bartle2016physics,innmann2016volumedeform, newcombe2015dynamicfusion}, hand-crafted priors~\cite{kumar2017monocular, ranftl2016dense, russell2014video}, low-rank assumptions for motion modeling~\cite{bregler2000recovering, dai2014simple, novotny2019c3dpo}, integrating monocular depth estimation priors~\cite{kopf2021robust, li2025megasam, zhang2022structure} or large-scale training data~\cite{lu2025align3r, wang2025continuous, wang2024dust3r, zhang2024monst3r, han2025d}. Recent works extend powerful representations such as Neural Radiance Field (NeRF)~\cite{mildenhall2021nerf} and 3D Gaussian Splatting (3DGS)~\cite{kerbl20233d} to dynamic scenarios~\cite{du2021neural, park2021hypernerf,fridovich2023k,luiten2024dynamic}. NeRF-based methods augment the network architecture with additional components such as flow fields~\cite{du2021neural, li2021neural, li2023dynibar, wang2021neural}, deformable neural fields~\cite{park2021nerfies, park2021hypernerf, pumarola2021d}, dynamic time conditioning~\cite{song2023nerfplayer}, or voxel-based representations~\cite{fridovich2023k, cao2023hexplane}. 3DGS-based methods further address NeRF's limited rendering quality and slow rendering speeds, incorporating dynamic designs such as deformable fields~\cite{luiten2024dynamic, yang2024deformable}, 4D Gaussian primitives~\cite{wu20244d},  velocity-integrated Gaussians~\cite{wang2025freetimegs}, or motion bases~\cite{lei2025mosca, wang2025shape}. While these methods achieve compelling 4D reconstruction results, they require per-scene optimization, making it difficult to generalize across multiple scenes.

\paragrapht{Feed-forward 4D Reconstruction.}
To address these limitations, recent approaches have explored generalizable feed-forward networks~\cite{xu20254dgt, liang2024feed, shen2025seeing,han2025d, ma20254d, lin2025dgs, hurufo, yang2026neoverse, wu2025streamsplat} by learning priors from large-scale datasets or leveraging foundational priors~\cite{leroy2024grounding, wang2025vggt}. These methods either generate Gaussians at a specific timestamp~\cite{shen2025seeing, liang2024feed} or additionally predict per-Gaussian velocities to capture motion~\cite{xu20254dgt, ma20254d, lin2025dgs, hurufo}. Despite their success, predicting per-pixel Gaussians yields millions of primitives, introducing redundancy, visual artifacts, and high computational cost. Moreover, such per-pixel modeling induces a view-dependent bias~\cite{an2025c3g, wang2025volsplat, miao2025evolsplat, kim20244d}: rendering at a given timestamp relies predominantly on Gaussians from the corresponding input view, and in videos with smooth motion, interpolating nearby per-pixel Gaussians already yields high rendering scores—providing little incentive to reason about long-range motion. We instead reconstruct scenes through a compact set of learnable query tokens, an architectural bottleneck that compels the model to aggregate temporal information globally.

\paragrapht{Towards Addressing View-dependent Bias.}
The tendency of reconstruction models to overfit to input views is a long-standing challenge in 3D reconstruction, observed in both per-scene optimization~\cite{deng2022depth, kim2022infonerf, niemeyer2022regnerf, li2024dngaussian} and feed-forward settings~\cite{hong2024pf3plat, huang2025no, charatan2024pixelsplat}. Since photometric reconstruction objectives alone provide insufficient signal for the model to learn globally consistent geometry, existing works incorporate geometric regularizations such as monocular depth priors~\cite{roessle2022dense, song2023darf, wang2023sparsenerf, li2024dngaussian}, opacity constraints~\cite{kim2022infonerf, yang2023freenerf}, novel view self-supervision~\cite{niemeyer2022regnerf, kwak2023geconerf, jung2023self}, or epipolar supervision~\cite{charatan2024pixelsplat, chen2024mvsplat, wang2024freesplat,hong2024unifying}. This challenge is especially pronounced in dynamic scenes, where moving regions inherently break multi-view consistency, necessitating expert geometric models~\cite{wang2025shape, wang2025freetimegs} or explicit motion constraints~\cite{wang2025shape, lei2025mosca}. On the architectural side, recent works show that reducing Gaussian redundancy itself mitigates view-dependent bias: EvolSplat~\cite{miao2025evolsplat} and VolSplat~\cite{wang2025volsplat} estimate Gaussians per voxel only when needed, while C3G~\cite{an2025c3g} predicts a compact set of Gaussians through learnable query tokens. 

\section{Methodology}
\label{sec:methods}
We propose \ours, a learnable query-based framework that takes dynamic monocular videos as input and produces a compact representation of dynamic scenes. We begin by describing our problem formulation~(\S~\ref{subsec:problem}), followed by our architecture design including 4D Gaussian reconstruction module and VDM refinement module~(\S~\ref{subsec:architecture}), analysis of emergent properties in our trained \ours~(\S~\ref{subsec:analysis}) and feature lifting decoder based on findings of analysis~(\S~\ref{subsec:feature}).

\subsection{Problem Setup}
\label{subsec:problem}
Given a monocular video $\mathcal{V} = \{ I_t \in \mathbb{R}^{H \times W \times 3} \}_{t=1}^T$ with $T$ frames of width $W$ and height $H$, and corresponding timestamps $\mathcal{T} = \{ t \in \mathbb{R} \}_{t=1}^T$, our goal is to build a feed-forward model that render high-quality novel views at arbitrary timestamps $t_b \in [1, T]$ \emph{without} any calibrated camera poses. Given a set of context frames $\mathcal{V}_c \subseteq \mathcal{V}$ along with their corresponding timestamps $\mathcal{T}_c \subseteq \mathcal{T}$, the model outputs a compact set of $N$ 3D Gaussians $\{ \mathbf{G}^i_{t_b} \}_{i=1}^N$ representing the scene at a target timestamp $t_b$. Each Gaussian is defined as $\mathbf{G}^i = \{ \mu^i, \sigma^i, \Sigma^i, c^i \}$, where $\mu^i \in \mathbb{R}^3$ denotes the mean position, $\sigma^i \in [0,1)$ the opacity, $\Sigma^i \in \mathbb{R}^{3 \times 3}$ the covariance matrix, and $c^i \in \mathbb{R}^{3(l+1)^2}$ the view-dependent color encoded via spherical harmonics~\cite{fridovich2022plenoxels} with $l$ levels. Iterating over $t_b \in \mathcal{T}$ yields a full dynamic reconstruction, enabling on-demand scene reconstruction at any moment within the temporal context.

\subsection{Compact 4D Representation with Gaussians}
\label{subsec:architecture}
\begin{wrapfigure}[15]{r}{0.6\textwidth}
    \centering
    \vspace{-13pt}
    \includegraphics[width=\linewidth]{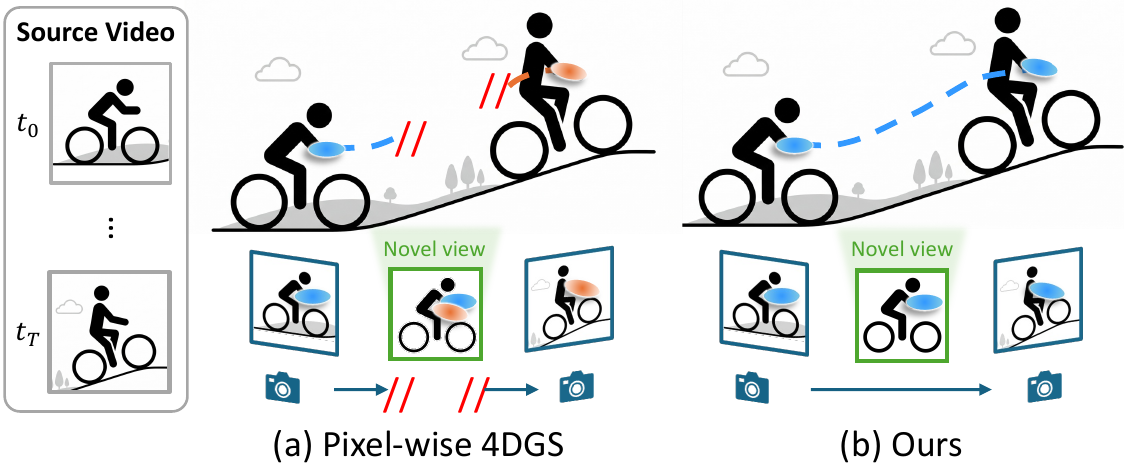}
    \vspace{-20pt}
    \caption{\textbf{Pixel-wise 4DGS vs. Ours.} \textbf{(a)} Pixel-wise methods produce duplicated, view-dependent Gaussians that cause ghosting at interpolated timestamps. \textbf{(b)} Our approach aggregate global temporal context, yielding a compact, unified Gaussian set with temporally coherent motion.
} 
    \label{fig:concept}
\end{wrapfigure}
As shown in Fig.~\ref{fig:concept}-(a), prior works~\cite{yang2026neoverse, xu20254dgt, lin2025movies} that estimate per-pixel Gaussians in dynamic settings produce redundant Gaussians tied to individual input views, where each Gaussian contributes reliably only near its observed timestamp. This leads to the model failing to capture the actual scene motion, where rendering at an interpolated timestamp results in duplicated Gaussians at inconsistent spatial locations---one set from each nearby input view---causing ghosting artifacts and degraded rendering quality. 

\begin{figure}[t]
    \centering
    \includegraphics[width=1\linewidth]{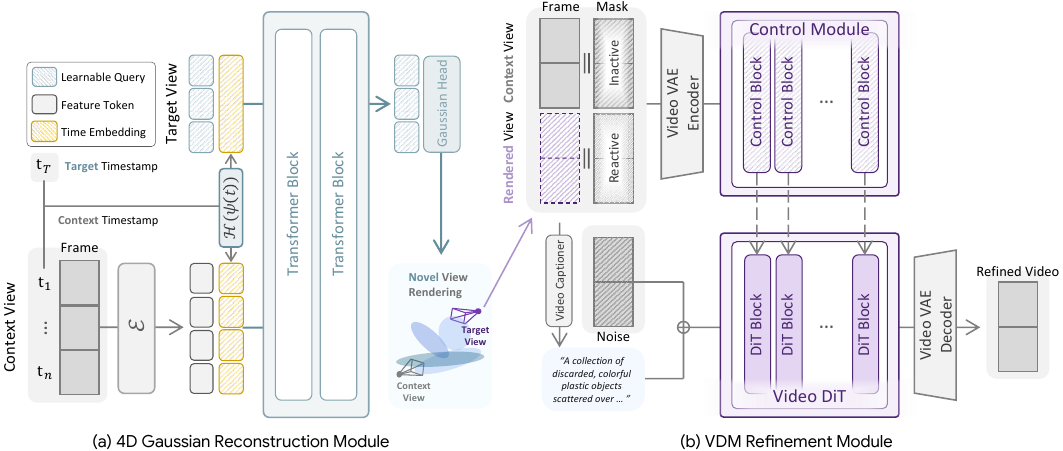}
    \vspace{-15pt}
    \caption{\textbf{Main architecture of \ours.} (a) A pre-trained encoder $\mathcal{E}$ extracts timestamp-injected features, which are decoded into 3D Gaussians by learnable query tokens conditioned on a target timestamp $t_b$. (b) A VDM refinement module that takes the rendered video as input and refines it conditioned on the context views.}
    \label{fig:architecture}
\end{figure}

To overcome this limitation, we propose a time-conditioned, query-based transformer decoder that employs a compact set of learnable queries to aggregate multi-frame information via full self-attention to better model the scene motion as shown in Fig.~\ref{fig:concept}-(b). Overall architecture is illustrated in Fig.~\ref{fig:architecture}.

\paragrapht{Visual Feature Extractor.}
To learn to decode compact set of Gaussians, we first extract geometry-grounded features $\mathbf{F}_{t} \in \mathbb{R}^{h \times w \times d}$ for each frame $I_{t}$ using an encoder $\mathcal{E}(\cdot)$, where $h$, $w$, and $d$ denote height, width and dimension of feature map. Following prior works~\cite{jiang2025anysplat}, we initialize our backbone from VGGT~\cite{wang2025vggt} weights, which provide strong geometry priors learned from large-scale data.

\paragrapht{Query-based Gaussian Decoder.}
Based on the extracted multi-view features, we adopt a query-based transformer decoder $\mathcal{D}_\mathcal{G}$ that introduces a limited set of learnable queries further decoded into Gaussians through interaction with extracted features $\mathbf{F}$ within the transformer. Specifically, we introduce a fixed number of $N$ learnable query tokens $\mathbf{Q} \in \mathbb{R}^{N \times d}$ which are then concatenated with visual features $[\mathbf{Q} ; \mathbf{F}]$, where $[ ; ]$ denotes concatenation. These combined features are passed through $L$ transformer layers with full self-attention, enabling bidirectional information interaction as follows:
\begin{equation}
    [\bar{\mathbf{Q}}; \bar{\mathbf{F}}] = \mathcal{D}_\mathcal{G} ([\mathbf{Q};\mathbf{F}]),
\end{equation}
where $\bar{\mathbf{Q}}$ and $\bar{\mathbf{F}}$ denote refined queries and features, respectively.
The refined query token $\bar{\mathbf{Q}}^i \in \mathbb{R}^d$ is then mapped to a 3D Gaussian $\mathbf{G}^i$ through a Gaussian head consisting of a single MLP layer.

\paragrapht{Time Embedding.}
To make the model temporally aware, we incorporate learnable timestamp embeddings into both the visual features and the learnable query tokens. First, we inject per-frame timestamp embeddings $t$ into the visual features of each context frame $\mathbf{F}_{t}$, enabling the model to distinguish the temporal origin of each input view. Then, to generate Gaussians at a desired target time moment $t_b$, we condition the learnable query tokens $\mathbf{Q}$ on the corresponding timestamp embedding $t_b$, making the decoded Gaussians aware of the rendering time they should represent:
\begin{align}
    \hat{\mathbf{F}}_{t} = \mathbf{F}_{t} + \mathcal{H}(\psi(t)), \quad \hat{\mathbf{Q}}_{t_b} = \mathbf{Q} + \mathcal{H}(\psi(t_b)),
\end{align}
where $\psi$ denotes the sinusoidal positional embedding and $\mathcal{H}$ denotes a two-layer MLP. These time-embedded features $\{\hat{\mathbf{F}}_t\}_{t=1}^T$ and queries $\hat{\mathbf{Q}}_{t_b}'$ are then fed into the Gaussian decoder.

This simple design offers two key benefits. First, the limited number of Gaussians forces the model to represent both static and dynamic regions with the same tokens, encouraging it to learn the underlying motion rather than memorize individual frames. Second, it enables efficient 4D reconstruction: by extracting visual features once and varying only the target timestamp $t_b$ across the lightweight transformer decoder, we decode Gaussians at arbitrary timestamps while reusing the same queries and features. As each query yields temporally smooth Gaussians under varying $t_b$, the dynamic scene is represented both compactly and efficiently.

\paragrapht{Rendering Enhancement.}
\label{subsec:vdm}
Keeping the number of Gaussians compact is essential for learning global motion, but inevitably trades off against rendering quality in high-frequency regions. Prior feed-forward 3DGS models~\cite{an2025c3g, jiang2025anysplat} recover fine-grained details via test-time optimization with 3DGS~\cite{kerbl20233d}, yet this is not directly applicable to dynamic scenes since 3DGS assumes static scenes with multi-view consistency. Meanwhile, recent works~\cite{lee20253d, seo2026grounding} leverage Video Diffusion Models (VDMs)~\cite{wan2025wan, jiang2025vace} to generate dynamic scenes while preserving static geometry, typically by warping point maps to novel viewpoints and relying on the diffusion model to fill holes and synthesize dynamic content.

In contrast, our method renders a geometrically coherent novel-view image that encompasses both static and dynamic components without producing disocclusion artifacts, allowing the VDM to focus solely on enhancing fine-grained scene details.
Specifically, we adopt Wan2.1-VACE-1.3B~\cite{jiang2025vace}, a diffusion model that learns to generate video conditioned on reference images following the ControlNet~\cite{zhang2023adding}. Given a context video $\mathcal{V}_c$ and rendered videos $\tilde{\mathcal{V}}_t$ from novel camera poses $\mathcal{P}_t$, we use the context video as a reference to provide the diffusion model with hints for refinement, while the rendered videos serve as the generation targets. Following \cite{jiang2025vace}, we assign a context mask $M_c = 0$ to the context frames (inactive) and a target mask $M_t = 1$ to the rendered frames (reactive) as control block inputs.
We encode reactive $\mathbf{z}_{\text{reactive}}$ and inactive latents $\mathbf{z}_{\text{inactive}}$ using VAE encoder $\mathcal{E}_\text{VAE}(\cdot)$:
\begin{align}
    \mathbf{z}_{\text{reactive}} &= [\mathcal{E}_\text{VAE}(\mathcal{V}_c \cdot M_c); \mathcal{E}_\text{VAE}(\tilde{\mathcal{V}}_t \cdot M_t)], \\
    \mathbf{z}_{\text{inactive}} &= [\mathcal{E}_\text{VAE}(\mathcal{V}_c \cdot (1-M_c)); \mathcal{E}_\text{VAE}(\tilde{\mathcal{V}}_t \cdot (1-M_t))].
\end{align}
We feed concatenation of inactive, reactive latents and masks into the control blocks, whose output features are added to the corresponding layers of the original VDM~\cite{wan2025wan}. After training, it serves as a post-processing step that refines \ours\ outputs through iterative denoising. Further details are in \S~\ref{supsubsec:diffusion}.

\paragrapht{Loss Functions.}
\label{subsec:loss}
The 4D Gaussian reconstruction module is supervised by a photometric loss $\mathcal{L}_{\text{color}}$, which combines MSE and LPIPS losses computed on target viewpoints. However, dynamic regions provide limited multi-view consistency cues, making it difficult to recover accurate geometry from monocular video alone. We therefore introduce two auxiliary supervision signals derived from foundation models: depth and normal losses ($\mathcal{L}_{\text{depth}}$, $\mathcal{L}_{\text{normal}}$) from MoGe-2~\cite{wang2025moge} to improve geometric fidelity, and a tracking loss $\mathcal{L}_{\text{track}}$ from CowTracker~\cite{lai2026cowtracker} to strengthen motion modeling. 

The VDM refinement module is conditioned on the VACE context features—extracted from the renderings of the 4D reconstruction module and the context images—together with a text prompt generated by Qwen3-VL-8B-Instruct~\cite{bai2025qwen3vl}. Given these conditions, the VDM generates refined renderings from noise and is trained with flow matching~\cite{lipman2022flow}. Further details are provided in \S~\ref{supsubsec:loss}.

\subsection{Analysis of Emergent Properties}
\label{subsec:analysis}
While trained solely with the rendering loss, we observe that introducing a compact set of query tokens enables our model to better capture global motion and scene geometry, in contrast to per-pixel methods that tend to overfit to individual input views. To investigate how this capability emerges, we examine how the learnable query tokens aggregate multi-frame features in dynamic scenes. Specifically, we visualize the attention maps inside the self-attention layers of $\mathcal{D}_\mathcal{G}$, where queries are taken from the time-conditioned learnable tokens $\hat{\mathbf{Q}}_{t_b}$ and keys from the visual features $\{\hat{\mathbf{F}}_t\}_{t=1}^T$.

\begin{figure}[t]
    \centering
    \includegraphics[width=\linewidth]{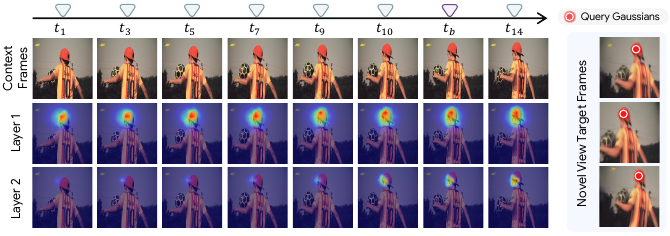}
    \vspace{-15pt}
    \caption{\textbf{Analysis of attention patterns.} Visualization of attention maps between the learnable query tokens and multi-frame image features. For the query token decoding a specific Gaussian (\textcolor{red}{red} dot), the two self-attention layers exhibit complementary behaviors: the first attends to geometrically corresponding regions across all frames, while the second concentrates on frames temporally close to the target timestamp.} 
    \label{fig:analysis}
    \vspace{-10pt}
\end{figure}

As shown in Fig.~\ref{fig:analysis}, our model exhibits a clear emergent property in its attention behavior, where the two self-attention layers play complementary roles. The first layer attends broadly to geometrically corresponding regions across all frames without explicit supervision, while the second focuses on frames temporally close to the target timestamp $t_b$. Together, they yield attention maps that are both spatially and temporally aware, allowing each query to retrieve only the features needed to decode its Gaussian. We believe this arises from an implicit optimization pressure: to reconstruct dynamic scenes at arbitrary timestamps with a limited number of Gaussians, the model is compelled to position them at geometrically coherent regions while reasoning about scene motion across frames.

\subsection{Any-feature 4D Lifting}
\label{subsec:feature}
Thanks to this emergent property, we observe that the learned attention maps of $\mathcal{D}_\mathcal{G}$ can be directly leveraged to lift arbitrary 2D vision foundation model (VFM) features into 4D in a feed-forward manner. Such 4D feature lifting is essential for a wide range of downstream applications, including dynamic 3D scene understanding and 4D correspondence. However, extending feed-forward 4DGS networks to support feature lifting has remained fundamentally difficult as their overfitting to input views prevents them from modeling global scene motion, which is an essential prerequisite for resolving which 2D feature corresponds to which 3D Gaussian across frames. As a result, existing 4D feature lifting approaches~\cite{zhou2025feature4x, ji2024segment} have been confined to per-scene optimization, relying on explicit backward warping along Gaussian trajectories and incurring substantial computational overhead. Leveraging the emergent properties of $\mathcal{D}_\mathcal{G}$, which already aggregates multi-frame features in a globally consistent manner, we overcome this limitation and present the first feed-forward feature-lifting architecture that directly generates 4D feature fields.

\paragrapht{View-invariant 4D Feature Decoder.}
Specifically, following~\cite{an2025c3g}, we design a feature decoder $\mathcal{D}_\mathcal{F}$ that mirrors the architecture of $\mathcal{D}_\mathcal{G}$ and reuses its attention patterns to aggregate features from any VFM. $\mathcal{D}_\mathcal{F}$ takes as input the features $\mathbf{F}'_t = \mathcal{E}'(I_t)$ extracted by an arbitrary vision encoder $\mathcal{E}'$ (e.g., DINOv3~\cite{simeoni2025dinov3} or VGGT~\cite{wang2025vggt}), together with the same learnable queries $\mathbf{Q}$ used in $\mathcal{D}_\mathcal{G}$:
\begin{equation}
    [\bar{\mathbf{Q}}'; \bar{\mathbf{F}}'] = \mathcal{D}_\mathcal{F} ([\mathbf{Q};\mathbf{F'}]),
\end{equation}
where $\bar{\mathbf{Q}}'$ denotes the queries refined by the feature decoder. To exploit the learned attention patterns, we reuse the queries and keys from $\mathcal{D}_\mathcal{G}$ and train only the value projection layers. The refined queries are then passed through a single-MLP head to produce the feature attribute of each Gaussian. Since the Gaussians and feature attributes are generated from the same learnable tokens, augmenting each Gaussian with its feature attribute naturally yields a 4D feature field. Further details are in \S~\ref{supsubsec:feature_lifting}.

\section{Experiments}
\label{sec:exp}
\subsection{Implementation Details}
\label{subsec:imde}
We set the number of learnable queries to $N=2048$ and using $L=2$ transformer layers, with VGGT~\cite{wang2025vggt} as the visual encoder $\mathcal{E}(\cdot)$. For the Gaussian representation, we adopt the standard 3DGS~\cite{kerbl20233d} formulation but fix the spherical harmonic degree to 0, representing appearance as a single RGB color per Gaussian. This choice suppresses view-directional biases and improves training stability.
For training, we use input images of resolution $224 \times 224$. We use Spring~\cite{mehl2023spring}, Kubric~\cite{greff2022kubric} and RealEstate10K~\cite{zhou2018stereo} datasets. The loss weights are set to $\lambda_{\text{depth}}=0.001$, $\lambda_{\text{normal}}=0.001$, and $\lambda_{\text{track}}=0.1$. We optimize the model using the AdamW optimizer~\cite{loshchilov2017decoupled} with separate learning rates of $1e^{-5}$ for the transformer decoder and $1e^{-7}$ for the visual backbone, combined with a cosine annealing scheduler that decays to 0.1 of the initial learning rate. Training is conducted with a per-GPU batch size of 1 across 4 NVIDIA H100 GPUs. To stabilize training, we initialize the model from C3G weights pretrained on static scenes, allowing the optimization to focus on the dynamic modeling. Further dataset and evaluation details are described in \S~\ref{supsubsec:dataset} and~\ref{supsec:evalpro}.

\begin{table}[t]
    \centering
    \caption{\textbf{Quantitative results of novel view synthesis on dynamic datasets.} Our model achieves superior rendering quality across all datasets without requiring GT poses, while using a limited number of Gaussians. \textbf{Bold} denotes the best and \underline{underline} denotes the secondary performance.} 
    \label{tab:novel_view_synthesis}
    \vspace{-5pt}
    \resizebox{\linewidth}{!}{
    \begin{tabular}{l|c|c|c|ccc|ccc|ccc|ccc}
        \toprule
        \multirow{2}{*}{Methods} & \multirow{2}{*}{\shortstack{Per \\ scene}} & \multirow{2}{*}{Pose} & \multirow{2}{*}{$ \# \mathbf{G} \downarrow$} &\multicolumn{3}{c|}{DyCheck~\cite{gao2022monocular}} & \multicolumn{3}{c|}{ADT~\cite{pan2023aria}} & \multicolumn{3}{c|}{TUM-Dynamics~\cite{sturm2012benchmark}} & \multicolumn{3}{c}{NVIDIA~\cite{yoon2020novel}} \\
        & & & & PSNR$\uparrow$ & SSIM$\uparrow$ & LPIPS$\downarrow$ & PSNR$\uparrow$ & SSIM$\uparrow$ & LPIPS$\downarrow$ & PSNR$\uparrow$ & SSIM$\uparrow$ & LPIPS$\downarrow$ & PSNR$\uparrow$ & SSIM$\uparrow$ & LPIPS$\downarrow$ \\
        \midrule \midrule
        Shape of Motion~\cite{wang2025shape} & \ding{51} & \ding{51} & 128K & \underline{14.13} & 0.313 & 0.690 & - & - & - & 14.29 & 0.453 & 0.552 & 10.86 & 0.280 & 0.791 \\
        MoSca~\cite{lei2025mosca} & \ding{51} & \ding{51} & 342K & 11.93 & 0.252 & 0.700 & 2.419 & 0.343 & 0.892 & 9.765 & 0.247 & 0.654 & 12.62 & 0.269 & 0.691 \\
        \midrule
        4DGT~\cite{xu20254dgt} & \ding{55} & \ding{51} & 272K & 12.15& 0.247& 0.477 & 19.22& \textbf{0.717}& \textbf{0.260}& \underline{17.27}& \underline{0.578}& \underline{0.329}& 15.64& \underline{0.349}&0.614  \\
        MoVieS~\cite{lin2025movies} &\ding{55} &\ding{51} & 802K & 11.99& \underline{0.352}& \textbf{0.359}& \underline{20.35}& 0.526& 0.394& 14.91& 0.378& 0.395& 13.71&0.211& 0.596 \\
        \midrule
        NeoVerse~\cite{yang2026neoverse}  & \ding{55} & \ding{55} & 802K& 11.90& 0.251& 0.450& {21.94}& 0.629& \underline{0.270}& 15.26& 0.437& 0.415& \underline{15.86}& 0.322&\underline{0.492} \\
        Ours & \ding{55} & \ding{55} & \textbf{2K} & \textbf{15.64}& \textbf{0.388}& \underline{0.384}& \textbf{22.35}& \underline{0.631}& 0.313& \textbf{19.52}& \textbf{0.603}& \textbf{0.306}& \textbf{20.51}& \textbf{0.489}&\textbf{0.393} \\
        \bottomrule
    \end{tabular}}
    \vspace{-10pt}
\end{table}
\begin{table}[t]
    \centering
    \caption{\textbf{Quantitative results of novel view synthesis with various interval $\mathrm{\Delta t}$ on TUM-Dynamics.} Across varying temporal gaps between input frames, our model consistently maintains high-quality novel-view synthesis. \textbf{Bold} denotes the best and \underline{underline} denotes the secondary performance.}
    \label{tab:novel_view_synthesis_gap}
    \vspace{-5pt}
    \resizebox{\linewidth}{!}{
    \begin{tabular}{l|c|ccc|ccc|ccc|ccc}
        \toprule
        \multirow{2}{*}{Methods} & \multirow{2}{*}{Pose} &\multicolumn{3}{c|}{$\mathrm{\Delta t=2}$} & \multicolumn{3}{c|}{$\mathrm{\Delta t=4}$} & \multicolumn{3}{c|}{$\mathrm{\Delta t=6}$} & \multicolumn{3}{c}{$\mathrm{\Delta t=8}$}\\
        & & PSNR$\uparrow$ & SSIM$\uparrow$ & LPIPS$\downarrow$ & PSNR$\uparrow$ & SSIM$\uparrow$ & LPIPS$\downarrow$ & PSNR$\uparrow$ & SSIM$\uparrow$ & LPIPS$\downarrow$ & PSNR$\uparrow$ & SSIM$\uparrow$ & LPIPS$\downarrow$ \\
        \midrule \midrule
        4DGT~\cite{xu20254dgt} & \ding{51} & 19.27& \textbf{0.674}& \underline{0.251}& 18.08& \textbf{0.623}& 0.297
& \underline{17.27}& \underline{0.578}& \underline{0.329}
    & \underline{16.27}& \underline{0.547}&\underline{0.351}\\
        MoVieS~\cite{lin2025movies} & \ding{51} & 16.30& 0.441& 0.314& 15.93& 0.432& 0.334& 14.91& 0.378& 0.395
& 14.23& 0.322&0.428\\
        \midrule
        NeoVerse~\cite{yang2026neoverse} & \ding{55} & \underline{20.14}& \underline{0.671} & \textbf{0.216}& \underline{18.23}& 0.575& \underline{0.296}& 15.26& 0.437& 0.415
& 15.94& 0.475&0.387\\
        Ours & \ding{55} & \textbf{20.59}& 0.647& 0.267& \textbf{20.00}& \underline{0.619}& \textbf{0.290}& \textbf{19.52}& \textbf{0.603}& \textbf{0.306}& \textbf{19.23}& \textbf{0.599}&\textbf{0.311}\\
        \bottomrule

    \end{tabular}}
    \vspace{-10pt}
\end{table}

\subsection{Evaluation Setting}
We evaluate novel-view synthesis performance across four dynamic datasets, each targeting a distinct aspect of dynamic scene reconstruction. \textbf{DyCheck}~\cite{gao2022monocular} captures each scene using one casually moving camera and two fixed cameras at synchronized time intervals. Following NeoVerse~\cite{yang2026neoverse}, we sample $T=64$ consecutive timestamps per scene, using 32 frames at stride 2 from the casually-captured video (camera 0) as context input $\mathcal V_c$ and all 64 frames from a fixed-camera video (camera 1) as evaluation targets. Metrics are computed exclusively over co-visible regions using co-visibility masks~\cite{wang2025shape}. This setting evaluates novel-view synthesis at \emph{observed} timestamps during input, assessing the model's ability to render unseen viewpoints at known times.
\textbf{ADT}~\cite{pan2023aria}, \textbf{TUM-Dynamics}~\cite{sturm2012benchmark}, and \textbf{NVIDIA}~\cite{yoon2020novel} each provide a single camera trajectory per scene. We select 16 temporally strided frames with stride $\mathrm{\Delta t}$ as context input and evaluate on mid-frames between consecutive context frames. By default, we set $\mathrm{\Delta t}=6$ for ADT and TUM-Dynamics, and $\mathrm{\Delta t}=2$ for NVIDIA. This setting evaluates novel-view synthesis at \emph{unobserved} timestamps during input, assessing the model's capacity for temporal interpolation and global motion understanding.


\subsection{Novel View Synthesis}

\begin{figure}[!t]
    \centering
    \includegraphics[width=\linewidth]{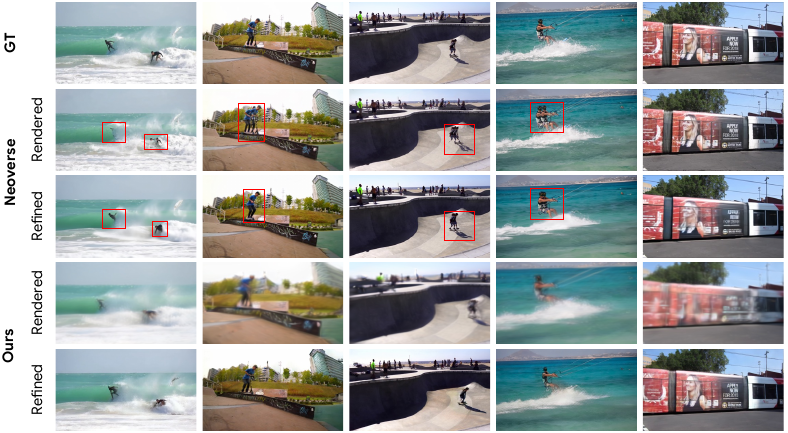}
    \vspace{-15pt}
    \caption{\textbf{Qualitative results of novel view synthesis on dynamic datasets.} We further provide qualitative comparisons between NeoVerse and \ours, showing both the rendered outputs of the feed-forward reconstruction model and the results after diffusion-based refinement. Our model exhibits fewer occlusion holes and ghost artifacts than NeoVerse, thereby mitigating hallucinations introduced by the diffusion-based enhancement module.
    }
    
    \label{fig:novel_view}
    \vspace{-10pt}
    
\end{figure}

We compare against per-scene optimization methods~\cite{wang2025shape,lei2025mosca} and feed-forward dynamic 3DGS networks, categorized into pose-required~\cite{xu20254dgt, lin2025movies} and pose-free~\cite{yang2026neoverse} settings. As shown in Tab.~\ref{tab:novel_view_synthesis} and Fig.~\ref{fig:novel_view}, \ours\ achieves superior rendering quality across all dynamic datasets \textbf{without requiring ground-truth poses}, using only \textbf{0.007}$\times$ or fewer Gaussians compared to competing methods.
Per-scene optimization methods tend to overfit to sparse context input views or fail to reconstruct entirely (\textit{e.g.}, Shape of Motion on the ADT dataset), resulting in degraded multi-view consistency and poor novel-view rendering quality. Feed-forward baselines suffer from per-view timestamp overfitting due to their pixel-wise prediction design, which limits their ability to capture global scene motion and introduces ghosting artifacts at novel viewpoints. 

In contrast, our model assigns each learnable query token to a distinct semantic region of the scene, enabling it to capture the full motion trajectory of that region across time. By aggregating information across all input views, this design encourages a more complete understanding of scene dynamics and constrains each Gaussian to occupy geometrically plausible locations even at unseen timestamps. 
This advantage is particularly evident on DyCheck~\cite{gao2022monocular}, where existing methods that model individual timestamps tend to rely exclusively on Gaussians from temporally nearby frames, suppressing temporally distant Gaussians by zeroing out their opacity $\sigma^i$, which introduces hole artifacts in the rendered output. In contrast, our model conditions Gaussian placement on the full global temporal context regardless of the target timestamp, enabling complete scene coverage and eliminating such artifacts. We further provide qualitative comparisons between NeoVerse and \ours\ in Fig.~\ref{fig:novel_view}, showing both the rendered outputs of the feed-forward reconstruction model and the results after diffusion-based refinement. As shown, our model produces fewer occlusion holes and ghost artifacts than NeoVerse, providing cleaner anchors to the diffusion module and thereby mitigating hallucinations during refinement.

We further evaluate the robustness of feed-forward models to temporally sparse inputs in Tab.~\ref{tab:novel_view_synthesis_gap}. Specifically, we vary the input stride $\mathrm{\Delta t} \in \{2, 4, 6, 8\}$ on the TUM-Dynamics~\cite{sturm2012benchmark} and measure novel-view synthesis quality. As shown, ours exhibits significantly less performance degradation as the temporal gap increases compared others. We attribute this to the inherent limitation of pixel-wise methods: each Gaussian represents only temporally nearby regions relative to its observed timestamp, making it difficult to bridge large inter-frame displacements that demand a broader understanding of object motion. In contrast, our model leverages learnable tokens to aggregate information across the full temporal context, enabling robust novel-view synthesis even under large temporal gaps.

\begin{table}[t]
    \centering
    \begin{minipage}{0.58\linewidth}
        \caption{\textbf{Temporal-invariant feature.} } 
        \label{tab:point_tracking}
        \vspace{-5pt}
        \centering
        \resizebox{\linewidth}{!}{
        \begin{tabular}{l|cccc|cccc}
            \toprule
                \multirow{2}{*}{Methods} & \multicolumn{4}{c|}{RGB-stacking~\cite{lee2021beyond, doersch2022tap}} & \multicolumn{4}{c}{DAVIS~\cite{pont20172017, doersch2022tap}} \\
                & $<\delta^0$ & $<\delta^2$ & $<\delta^4$ & $<\delta^z_\text{avg}$ & $<\delta^0$ & $<\delta^2$ & $<\delta^4$ & $<\delta^z_\text{avg}$ \\
                \midrule \midrule
                DINOv3-L~\cite{simeoni2025dinov3} & 1.3 & 11.3 & 65.6 & 23.2 & 0.6& 6.2& 40.7& 13.7\\
                DINOv3-L~\cite{simeoni2025dinov3} + Ours & \textbf{6.9}& \textbf{38.3}& \textbf{74.4}& \textbf{39.3}& \textbf{2.2}& \textbf{19.8}& \textbf{63.1}& \textbf{27.1}\\
            \midrule
                VGGT-Tracking~\cite{wang2025vggt} & 1.1& 15.9& 63.6& 24.8& 2.7& 17.9& 54.0& 23.1\\
                VGGT-Tracking~\cite{wang2025vggt} + Ours & \textbf{4.9}& \textbf{28.7}& \textbf{68.0}& \textbf{33.0}& \textbf{4.4}& \textbf{23.4}& \textbf{68.7}& \textbf{30.5}\\
            \bottomrule
        \end{tabular}}
    \end{minipage}
    \hfill
    \begin{minipage}{0.4\linewidth}
        \caption{\textbf{Dynamic scene understanding}}
        \label{tab:scene_understanding}
        \vspace{-5pt}
        \centering
        \resizebox{\linewidth}{!}{
        \begin{tabular}{l|ccccc}
            \toprule
                \multirow{2}{*}{Methods} & \multicolumn{5}{c}{DAVIS~\cite{pont20172017}} \\
                & mIOU$\uparrow$ & Acc$\uparrow$ & PSNR$\uparrow$ & SSIM$\uparrow$ & LPIPS$\downarrow$ \\
                \midrule \midrule
                LSeg~\cite{li2022language} & 0.550 & 0.801 & -& -& -\\
                \midrule
                LSM~\cite{fan2024large} & 0.615& 0.844& 14.320& 0.310& 0.517\\
                C3G~\cite{an2025c3g} & 0.345 & 0.528 & 18.623 & 0.376 & 0.460 \\
                Ours & \textbf{0.634} & \textbf{0.854} & \textbf{20.181} & \textbf{0.426} & \textbf{0.359} \\
            \bottomrule
        \end{tabular}}
    \end{minipage}
    \vspace{-10pt}
\end{table}

\subsection{Feature Lifting}
\paragrapht{Temporal-invariant Feature Field.}
To validate the effectiveness of our 4D feature field, we conduct experiments on point tracking and scene understanding, following the evaluation protocol of TAP-Vid~\cite{doersch2022tap}. Specifically, we extract per-frame features and perform nearest-neighbor matching based on feature similarity. A well-constructed 4D feature field should yield projected features that are more temporally and geometrically consistent than the original 2D features. As shown in Tab.~\ref{tab:point_tracking}, features projected from our 4D field outperform the original vision foundation model features (\textit{e.g.}, DINOv3~\cite{simeoni2025dinov3} and VGGT~\cite{wang2025vggt}), demonstrating that our model produces a 4D feature field with improved geometric and temporal consistency.

\paragrapht{Dynamic Scene Understanding.}
For scene understanding, our framework can lift arbitrary 2D features into the Gaussian field, allowing it to leverage semantically rich foundation models such as CLIP~\cite{radford2021learning} and LSeg~\cite{li2022language}. Each Gaussian is thus augmented with a semantic embedding that remains consistent across both viewpoints and timestamps, enabling open-vocabulary reasoning~\cite{kim2026seg4diff,shin2024towards,cho2024cat} directly in the 4D space. We evaluate this on the DAVIS dataset~\cite{pont20172017}, using a subset of sequences in which LSeg can reliably distinguish foreground objects. As shown in Tab.~\ref{tab:scene_understanding}, our model significantly outperforms prior feed-forward scene understanding methods, which are largely restricted to static scenes. Moreover, by aggregating features across multiple frames, our model also surpasses LSeg itself, demonstrating that the proposed 4D feature field not only inherits the semantic priors of foundation models but further enhances them through temporally coherent aggregation.

Together, these results demonstrate that our 4D feature field offers a compact yet expressive representation that maintains consistency across both spatial and temporal axes. Furthermore, by lifting semantically rich features into each Gaussian, the field naturally augments per-Gaussian semantic information, enabling spatially and temporally coherent dynamic scene understanding.

\subsection{Ablation Studies}
\label{subsec:ablation}
We conduct ablation studies on the time embedding and loss function designs to validate the contribution of each component. In Tab.~\ref{tab:ablation_time}, we ablate the choice of time embedding used to condition the learnable query tokens on the target timestamp. The results show that sinusoidal embeddings consistently outperform RoPE embeddings, and increasing the embedding dimensionality yields steady improvements up to 256, beyond which returns saturate.

In Tab.~\ref{tab:ablation_loss}, we ablate the contribution of each loss term. The results demonstrate that expert-guided supervision signals, specifically depth $\mathcal L_{\text{depth}}$, normal $\mathcal L_{\text{normal}}$ and tracking $\mathcal L_{\text{track}}$, substantially improves performance over training with photometric loss $\mathcal L_{\text{color}}$ alone, confirming that these auxiliary losses are critical for alleviating the geometric ambiguity inherent in monocular dynamic reconstruction.

        

\begin{table}[t]
    \centering
    \begin{minipage}{0.49\linewidth}
        \centering
        \caption{\textbf{Ablation studies on time embedding.}}
        \label{tab:ablation_time} 
        \vspace{-5pt}
        \resizebox{\linewidth}{!}{
        \begin{tabular}{c|c|ccc|ccc}
            \toprule
            \multirow{2}{*}{\shortstack{Positional \\ Embeddings}} & \multirow{2}{*}{Dim.} & \multicolumn{3}{c|}{DyCheck~\cite{gao2022monocular}} & \multicolumn{3}{c}{TUM-Dynamics~\cite{sturm2012benchmark}} \\
             & & PSNR $\uparrow$ & SSIM $\uparrow$ & LPIPS $\downarrow$ & PSNR $\uparrow$ & SSIM $\uparrow$ & LPIPS $\downarrow$ \\
            \midrule \midrule
            RoPE & - & 12.73 & 0.302 & 0.520 & 14.90 & 0.349 & 0.571 \\
            Sinusoidal & 64 & 15.31 & 0.382 & 0.397 & 18.65 & 0.559 & 0.363\\
            Sinusoidal & 128 & 15.39 & 0.377 & 0.388 & 18.67 & 0.560 & 0.355\\
            \textbf{Sinusoidal} & \textbf{256}& \textbf{15.54} & 0.383 & \textbf{0.385} & \textbf{18.94} & \textbf{0.575} & \textbf{0.341}\\
            Sinusoidal & 512 & \textbf{15.54} & \textbf{0.386} & 0.391 & 18.90 & 0.572 & 0.345\\
            
            \bottomrule
        \end{tabular}}
    \end{minipage}
    \hfill
    \begin{minipage}{0.49\linewidth}
        \centering
        \caption{\textbf{Ablation studies on loss function.}}
        \label{tab:ablation_loss}
        \vspace{-5pt}
        \resizebox{\linewidth}{!}{
        \begin{tabular}{l|ccc|ccc}
            \toprule
            \multirow{2}{*}{Methods} & \multicolumn{3}{c|}{DyCheck~\cite{gao2022monocular}} & \multicolumn{3}{c}{TUM-Dynamics~\cite{sturm2012benchmark}} \\
             & PSNR $\uparrow$ & SSIM $\uparrow$ & LPIPS $\downarrow$ & PSNR $\uparrow$ & SSIM $\uparrow$ & LPIPS $\downarrow$ \\
            \midrule \midrule
            Ours & \textbf{15.54} & 0.383 & \textbf{0.385} & 18.94 & \textbf{0.575} & \textbf{0.341}\\
            w/o $\mathcal{L}_{\text{track}}$ & 15.46& 0.381& 0.390& \textbf{18.97}& 0.574& 0.342\\
            w/o $\mathcal{L}_\text{depth}$ & 15.50& 0.381& \textbf{0.385}& 18.35& 0.557& 0.357\\
            w/o $\mathcal{L}_\text{normal}$ & 15.41& \textbf{0.390}& 0.389& 17.86& 0.537& 0.375\\
            w/o $\mathcal{L}_{\text{depth}}, \mathcal{L}_{\text{normal}}$ & 15.21& 0.377& 0.397& 18.80& 0.568& 0.347\\
            
            \bottomrule
    \end{tabular}}
        
    \end{minipage}\vspace{-10pt}
\end{table}
\section{Conclusion}
\label{sec:conclusion}
In this work, we present \ours, a feed-forward framework for dynamic scene reconstruction from monocular video that departs from pixel-wise Gaussian prediction. By introducing a compact set of timestamp-conditioned learnable query tokens, \ours{} aggregates multi-frame features across the full temporal context and decodes 3D Gaussians whose positions are adaptively modulated by the target timestamp. This globally motion-aware Gaussian representation further supports a range of downstream applications, including VDM-based rendering enhancement, point tracking, and 4D feature lifting. Across multiple benchmarks, \ours{} consistently outperforms existing feed-forward dynamic 3DGS methods—without requiring ground-truth poses, using significantly fewer Gaussians, and demonstrating particular robustness to large temporal gaps.

\newpage
\bibliographystyle{splncs04}
\bibliography{main}


\clearpage
\appendix

\title{Learning Global Motion with Compact Gaussians for Feed-Forward 4D Reconstruction \\ \texttt{--Supplementary Material--}}

\makeatletter
\begingroup
  \vbox{%
    \hsize\textwidth
    \linewidth\hsize
    \vskip 0.1in
    \@toptitlebar
    \centering
    {\LARGE\bf \@title\par}
    \@bottomtitlebar
    \vskip 0.3in \@minus 0.1in
  }
\endgroup
\makeatother


This supplementary material provides additional implementation details and experimental results that could not be included in the main paper due to space constraints. The supplementary is organized as follows.

Sec.~\ref{supsec:imde} introduces additional implementation details covering architecture, loss functions, and dataset curation and sampling strategies. Sec.~\ref{supsec:evalpro} provides details on the evaluation datasets and the comparison protocol used for feed-forward 4D Gaussian methods. Sec.~\ref{supsec:exp} presents additional experiments and ablation studies. Sec.~\ref{supsec:tracking} reports additional quantitative and qualitative results on point tracking. Sec.~\ref{supsec:addqual} provides further qualitative results. Finally, Sec.~\ref{supsec:limitation} discusses the limitations of our model.

We additionally provide \textbf{video results} on the DAVIS dataset~\cite{pont20172017}, a challenging in-the-wild video benchmark. Since DAVIS does not provide ground-truth camera poses, we present qualitative comparisons exclusively between \ours\ and NeoVerse~\cite{yang2026neoverse}, as both are pose-free feed-forward 4D Gaussian Splatting frameworks that do not require camera pose inputs at inference time.

\section{Additional Implementation Details}
\label{supsec:imde}
\subsection{Architecture Details}
\paragrapht{Visual Feature Extractor.}
Given a monocular video $\mathcal{V} = \{I_t \in \mathbb{R}^{H \times W \times 3} \}_{t=1}^T$ with corresponding timestamps $\mathcal{T} = \{t \in \mathbb{R} \}_{t=1}^T$ where $H$, $W$ and $T$ denote the frame height, width, and total number of frames respectively, we sample a subset of context frames $\mathcal{V}_c = \{I_t \}_{t\in \mathcal{T}_c}$ where $\mathcal{T}_c \subseteq \mathcal{T}$ denotes the sampled frame indices.
Following prior works~\cite{charatan2024pixelsplat, an2025c3g}, we extract visual features $\{ \mathbf{F}_t \in \mathbb{R}^{h \times w \times d} \}_{t \in \mathcal{T}_c}$ where $h$, $w$, and $d$ denote the spatial height, width and channel dimension respectively, using a pretrained geometry-grounded visual encoder $\mathcal{E}(\cdot)$ (\textit{e.g.}, VGGT~\cite{wang2025vggt}):
\begin{equation}
    \{ \mathbf{F}_t \}_{t \in \mathcal{T}_c} = \mathcal{E} ( \{ I_t \}_{t \in \mathcal{T}_c}).
\end{equation}

\paragrapht{Temporal Embedding.}
Given the extracted multi-view features $\{ \mathbf{F}_t \}_{t \in \mathcal{T}_c}$, we introduce $N$ learnable Gaussian queries $\mathbf{Q} \in \mathbb{R}^{N \times d}$, jointly optimized with the decoder. Each query is intended to decode a single 3D Gaussian representing the scene at the target time $t_b$.

To enable the model to distinguish representations across different timesteps and to condition the output on a specific target moment, we inject temporal embeddings into both the visual features and the learnable queries. We first normalize all timestamps relative to the target time $t_b$ and the earliest context frame:
\begin{equation}
    \hat{t} = \frac{t- \min(\mathcal{T}_c)}{t_b - \min(\mathcal{T}_c)}, \quad 
    \hat{t}_b = \frac{t_b- \min(\mathcal{T}_c)}{t_b - \min(\mathcal{T}_c)} = 1, 
\end{equation}

such that the target timestamp maps to 1 and all context timestamps are expressed as relative offsets. This normalization makes the model invariant to absolute time scales and encourages generalization across videos with varying temporal spans.

The normalized timestamps are encoded via sinusoidal positional embeddings $\psi(\cdot)$ and projected through a two-layer MLP $\mathcal{H}(\cdot)$, then added to their respective features:
\begin{align}
    \hat{\mathbf{F}}_{t} = \mathbf{F}_{t} + \mathcal{H}(\psi(\hat{t})), \quad \hat{\mathbf{Q}}_{t_b} = \mathbf{Q} + \mathcal{H}(\psi(\hat{t}_b)).
\end{align}

\paragrapht{Transformer Decoding.}
The time-conditioned visual features and queries are concatenated along the token dimension and passed through an $L$-layer Transformer decoder $\mathcal{D}(\cdot)$:
\begin{equation}
    [\bar{\mathbf{F}}_t;\bar{\mathbf{Q}}_{t_b}] = \mathcal{D}_\mathcal{G}([\hat{\mathbf{F}}_t;\hat{\mathbf{Q}}_{t_b}]),
\end{equation}
where $\bar{\mathbf{F}}_t \in \mathbb{R}^{h \times w \times d}$ and $\bar{\mathbf{Q}}_{t_b} \in \mathbb{R}^{N \times d}$ are the refined visual tokens and decoded query tokens respectively. The visual feature tokens serve as rich contextual keys and values, while the Gaussian queries attend over them to aggregate spatiotemporal scene information.

\paragrapht{Gaussian Head.}
Decoded query token $\bar{\mathbf{Q}}_{t_b}$ is independently mapped to a 3D Gaussians $\{\mathbf{G}^i_{t_b}\}_{i=1}^N$ via a single-layer MLP $\mathcal{M}_\mathcal{G}(\cdot)$:
\begin{equation}
    \{\mathbf{G}^i_{t_b}\}_{i=1}^N = \mathcal{M}_\mathcal{G}(\bar{\mathbf{Q}}_{t_b}).
\end{equation}
The resulting set of $N$ Gaussians $\{\mathbf{G}^i_{t_b}\}_{i=1}^N$ collectively represents the dynamic scene at target time $t_b$, and is rendered via differentiable Gaussians splatting. 

\subsection{Loss Details}
\label{supsubsec:loss}
\paragrapht{Photometric Loss.}
To propagate the learning signal, we render the predicted Gaussians onto target views $I_t \in \mathcal{V}$, which include context frames with known camera poses $\pi_t$. For each target timestamp $t$, we first estimate the Gaussians $\{\mathbf{G}_t^i \}_{i=1}^N$ using \ours, then render each pixel $p$ of the target view $\hat{I}_t$ through alpha blending of Gaussian color according to their depth order~\cite{kerbl20233d}:
\begin{equation}
    \hat{I}_t(p) = \sum_{i=1}^{N}c^i \sigma^i \mathbf{G}_t^{i,\text{2D}} \prod_{j=1}^{i-1}(1-\sigma^j \mathbf{G}_t^{j,\text{2D}}(p)),
\end{equation}
where $c^i$ is the view-dependent color of the $i$-th Gaussian obtained by decoding spherical harmonics coefficients at the viewing direction, and $\mathbf{G}_t^{i,\text{2D}}$ denotes the 2D screen-space projection of the 3D Gaussian under viewpoint $\pi_t$.

The photometric loss combines a pixel-wise mean squared error $\mathcal{L}_{\text{MSE}}$ with LPIPS loss $\mathcal{L}_{\text{LPIPS}}$ between rendered image $\hat{I}_t$ and ground-truth images $I_t$:
\begin{equation}
    \mathcal{L}_\text{color}(t) = \mathcal{L}_{\text{MSE}}(\hat{I}_t, I_t) + \lambda_{\text{LPIPS}} \mathcal{L}_{\text{LPIPS}}(\hat{I}_t, I_t),
\end{equation}
where $\lambda_{\text{LPIPS}}$ is the weighting coefficient for the LPIPS loss term, set to $0.05$ by default.

\paragrapht{Depth Loss.}
To encourage geometrically accurate Gaussian placement, we supervise the rendered depth map against pseudo ground-truth depths. The depth map $\hat{D}_t$ is rendered analogously to the color image via alpha-compositing, replacing color attributes with the $z$-values of the Gaussian means:
\begin{equation}
    \hat{D}_t(p) = \sum_{i=1}^{N}z^i \sigma^i \mathbf{G}_t^{i,\text{2D}} \prod_{j=1}^{i-1}(1-\sigma^j \mathbf{G}_t^{j,\text{2D}}(p)),
\end{equation}
where $z^i$ is the $z$-value of the mean position of Gaussians $\mathbf{G}^i_t$ in the camera coordinate frame. 

Since no ground-truth depth is available, we generate pseudo ground-truth depth maps $D_t$ using MoGe-2~\cite{wang2025moge}. As monocular depth predictions suffer from inherent scale-ambiguity, we adopt a scale-shift invariant loss~\cite{song2023darf, li2024dngaussian} rather than $l_1$ loss:
\begin{equation}
    \mathcal{L}_{\text{depth}}(t) = \|(\alpha \tilde{D}_{t} + \beta) - D_{t} \|,
\end{equation}
where $\alpha$ and $\beta$ are scale and shift parameters estimated via least-squares alignment~\cite{wang2025moge} between the rendered and pseudo ground-truth depth as follow:
\begin{equation}
    \alpha, \beta = \underset{\alpha,\beta}{\text{argmin}}\, \mathcal{L}_{\text{depth}}(t).
\end{equation}

\paragrapht{Normal Loss.}
Complementary to depth supervision, we additionally supervise surface normals to enforce local geometric consistency. The pseudo ground-truth normal map is generated from MoGe-2~\cite{wang2025moge}. Since the normal vectors are free from scale ambiguity, we directly apply an $l_1$ loss without affine alignment. 

We derive the rendered normal map $\hat{N}_t$ from the rendered depth map $\hat{D}_t$ by first back-projecting pixels into 3D using camera intrinsics $K$:
\begin{equation}
    \mathbf{P}(u,v) = \tilde{D}(u,v) K^{-1}[u, v, 1]^\intercal,
\end{equation}
where $(u,v)$ denotes the pixel coordinates. Surface normals are then estimated via the cross-product of finite-difference gradients:
\begin{equation}
    \hat{N}(u,v) = \mathrm{Norm}(\frac{\partial \mathbf{P}}{\partial u} \times \frac{\partial \mathbf{P}}{\partial v}).
\end{equation}
The normal loss $\mathcal{L}_{\text{normal}}$ is computed as:
\begin{equation}
    \mathcal{L}_{\text{normal}} = \| \hat{N} - N \|.
\end{equation}

\paragrapht{Tracking Loss.}
To suppress temporal flickering and Gaussian drift across frames, we introduce a tracking regularization loss that anchors Gaussian trajectories to predictions from a foundational point tracker.
For each timestamp $t \in \mathcal{T}$, we project the mean position $\mu^t$ of each Gaussian $\mathbf{G}_t$ onto the image plane using known extrinsics $\pi_t$ and intrinsics $K$:
\begin{equation}
    \mu_t^{\text{2D}} = K \pi_t \mu_t.
\end{equation}
We then randomly select a reference timestamp $t' \in \mathcal{T}_c$ and forward the corresponding projected points $\mu_{t'}^{\text{2D}}$ to a pretrained point tracker~\cite{lai2026cowtracker} $\mathcal{P}(\cdot,\cdot)$, which propagates them across the full video to yield dense 2D trajectories:
\begin{equation}
    \{\mathbf{p}_t\}^{T}_{t=1} = \mathcal{P}(\mathcal{V}, \mu^{\text{2D}}_{t'}).
\end{equation}

These predicted trajectories serve as soft constraints on the temporal motion of Gaussian means. We compute the tracking loss using a Huber penalty, masked by predicted point visibility:
\begin{equation}
    \mathcal{L}_{\text{track}} = \frac{1}{\sum_t m_{t}} \sum_t \left| m_{t} \cdot \text{Huber}(\| \mu^{\text{2D}}_{t} - \mathbf{p}_{t} \|) \right|,
\end{equation}
where $m_t$ is a visibility mask active when the predicted visibility score exceeds $0.5$.

\subsection{Feature Lifting Details}
\label{supsubsec:feature_lifting}
\paragrapht{Architecture details.}
For feature lifting, we first extract VFM features $\{ \mathbf{F}'_t \}_{t \in \mathcal{T}_c}$ from the context frames $\mathcal{V}_c = \{I_t\}_{t \in \mathcal{T}_c}$ using a VFM encoder $\mathcal{E}'(\cdot)$:
\begin{equation}
    \{ \mathbf{F}'_t \}_{t \in \mathcal{T}_c} = \mathcal{E}'(\{ I_t \}_{t \in \mathcal{T}_c}).
\end{equation}
Following the design of the \ours\ Gaussian decoder, we reuse the same $N$ learnable queries $\mathbf{Q} \in \mathbb{R}^{N \times d}$ from $\mathcal{D}_\mathcal{G}$, where each query is responsible for decoding the feature attribute of one 3D Gaussian at the target timestamp $t_b$. As in $\mathcal{D}_\mathcal{G}$, the normalized timestamps are encoded via sinusoidal positional embeddings $\psi(\cdot)$, projected through a two-layer MLP $\mathcal{H}(\cdot)$, and added to the corresponding features and queries:
\begin{align}
    \hat{\mathbf{F}}'_{t} = \mathbf{F}'_{t} + \mathcal{H}(\psi(\hat{t})), \quad \hat{\mathbf{Q}}'_{t_b} = \mathbf{Q}' + \mathcal{H}(\psi(\hat{t}_b)).
\end{align}

We then concatenate the VFM features and learnable queries and feed them into the feature decoder $\mathcal{D}_\mathcal{F}$, which shares the same architecture as $\mathcal{D}_\mathcal{G}$:
\begin{equation}
    [\bar{\mathbf{F}}'_t;\bar{\mathbf{Q}}'_{t_b}] = \mathcal{D}_\mathcal{F}([\hat{\mathbf{F}}'_t;\hat{\mathbf{Q}}'_{t_b}]),
\end{equation}
where $\bar{\mathbf{F}}'_t \in \mathbb{R}^{h \times w \times d}$ and $\bar{\mathbf{Q}}'_{t_b} \in \mathbb{R}^{N \times d}$ denote the refined feature tokens and decoded query tokens, respectively. Crucially, we reuse the query and key features $Q$ and $K$ from $\mathcal{D}_\mathcal{G}$ in every self-attention layer and propagate gradients only through the value projection. The attention is therefore computed as:
\begin{equation}
    \mathsf{Attn}(Q,K,V') = \mathsf{Softmax}\!\left(\frac{QK^\intercal}{\sqrt{d_Q}}\right) V',
\end{equation}
where $d_Q$ denotes the dimension of the query features, and $V'$ denotes the value features of $\mathcal{D}_\mathcal{F}$, obtained by projecting $[\hat{\mathbf{F}}'_t;\hat{\mathbf{Q}}'_{t_b}]$.

Decoded query token $\bar{\mathbf{Q}}'_{t_b}$ is independently mapped to a 3D Gaussians $\{\mathbf{f}^i\}_{i=1}^N$ via a single-layer MLP $\mathcal{M}_\mathcal{F}(\cdot)$:
\begin{equation}
    \{\mathbf{f}^i_{t_b}\}_{i=1}^N = \mathcal{M}_\mathcal{F}(\bar{\mathbf{Q}}'_{t_b}).
\end{equation}
The resulting set of $N$ feature attribute of each corresponding Gaussian $\{\mathbf{f}^i_{t_b}\}_{i=1}^N$ collectively represents the dynamic scene at target time $t_b$, and is rendered via differentiable Gaussians splatting.

\paragrapht{Loss function.}
Similar to color loss function, we render the predicted Gaussians onto target view $I_t \in \mathcal{V}$, which include context frames with known camera poses $\pi_t$. For each target timestamp $t$, we first estimate the Gaussians $\{\mathbf{G_t^i}\}_{i=1}^N$ and feature attribute $\{\mathbf{f_t^i}\}_{i=1}^N$ using \ours, then render each pixel $p$ of the target view through alpha blending of features which replace the color attributes of color rendering:
\begin{equation}
    \hat{\mathbf{F}}'_t(p) = \sum_{i=1}^{N}\mathbf{f}^i \sigma^i \mathbf{G}_t^{i,\text{2D}} \prod_{j=1}^{i-1}(1-\sigma^j \mathbf{G}_t^{j,\text{2D}}(p)).
\end{equation}
The feature rendering loss combines a L1 loss between projected feature $\hat{\mathbf{F}}'_t$ and pseudo ground truth $\mathbf{F}'_t$ which is extracted by VFM encoder $\mathbf{F}'_t = \mathcal{E}(I_t)$ given target view image $I_t$ as input:
\begin{equation}
    \mathcal{L}_\text{feat}(t) = \| \hat{\mathbf{F}}'_t - \mathbf{F}'_t  \|.
\end{equation}

\paragrapht{Implementation details.}
Building on the trained \ours\ weights, we additionally attach a feature decoder $\mathcal{D}_\mathcal{F}$ that shares the same architecture as $\mathcal{D}_\mathcal{G}$. To demonstrate generalizability across VFMs, we experiment with DINOv3~\cite{simeoni2025dinov3}, VGGT~\cite{wang2025vggt}, and LSeg~\cite{li2022language}. We adopt the same hyperparameters as \ours, except that we train $\mathcal{D}_\mathcal{F}$ for only 5K steps, since the attention patterns are already well-suited for aggregating relevant 2D features.

\subsection{VDM-based Rendering Enhancement Module Details}
\label{supsubsec:diffusion}
\paragrapht{Architecture details.}
The context video $\mathcal{V}_c \in \mathbb{R}^{H \times W \times T_c \times 3}$ with $T_c$ frames and the rendered video $\tilde{\mathcal{V}}_t \in \mathbb{R}^{H \times W \times T_t \times 3}$ with $T_t$ frames from \ours\ are given as input, where $H$ and $W$ denote the height and width, respectively. Following VACE~\cite{jiang2025vace}, our VDM-based rendering enhancement module consists of the original video DiT~\cite{wan2025wan} and a control module whose architecture follows ControlNet~\cite{zhang2023adding}. The original video DiT takes as input a noise tensor $\mathbf{z}_1 \sim \mathcal{N}(0,1)$ together with a caption $\mathcal{C}$ generated from the context video by Qwen3-VL-8B-Instruct~\cite{bai2025qwen3vl}.

For the control module, we adopt the VACE scheme, which designates the \emph{inactive} component as the reference frames and the \emph{reactive} component as the re-generated target frames. To enhance rendering quality conditioned on the context video, we treat the rendered video as reactive and the context video as inactive, setting the rendered-video active mask to $M_t=1$ and the context-video active mask to $M_c=0$. The reactive and inactive latents, $\mathbf{z}_{\text{reactive}}$ and $\mathbf{z}_{\text{inactive}}$, are then obtained through the VAE encoder $\mathcal{E}_{\text{VAE}}(\cdot)$ as follows:
\begin{align}
    \mathbf{z}_{\text{reactive}} &= [\mathcal{E}_\text{VAE}(\mathcal{V}_c \cdot M_c);\, \mathcal{E}_\text{VAE}(\tilde{\mathcal{V}}_t \cdot M_t)], \\
    \mathbf{z}_{\text{inactive}} &= [\mathcal{E}_\text{VAE}(\mathcal{V}_c \cdot (1-M_c));\, \mathcal{E}_\text{VAE}(\tilde{\mathcal{V}}_t \cdot (1-M_t))].
\end{align}

These latents are then concatenated along the temporal axis as $[\mathbf{z}_{\text{reactive}}; \mathbf{z}_{\text{inactive}}]$ and forwarded to the control module $\mathcal{R}_\text{control}$ as follows:
\begin{equation}
    \{\mathbf{f}_l\}_{l=0}^{L_\mathcal{R}} = \mathcal{R}_\text{control}([\mathbf{z}_{\text{reactive}}; \mathbf{z}_{\text{inactive}}]),
\end{equation}
where $L_\mathcal{R}$ denotes the number of layers in the control module and $\mathbf{f}_l$ denotes the hidden representation at layer $l$. These control features are then added to the corresponding hidden states of the original video DiT $\mathcal{R}_\text{VDM}$:
\begin{equation}
    \tilde{v}_{t'} = \mathcal{R}_\text{VDM}\bigl(\mathbf{z}_t,\, t',\, \{\mathbf{f}_l\}_{l=0}^{L_\mathcal{R}},\, \mathcal{C}\bigr),
\end{equation}
where $\mathbf{z}_t$ denotes intermediate latents at timestamp $t$, $\tilde{v}_{t'}$ denotes the estimated velocity at diffusion timestep $t'$, and $\mathcal{C}$ denotes the text caption condition.

\paragrapht{Loss function.}
For loss term, we use flow matching~\cite{lipman2022flow} loss following~\cite{wan2025wan, jiang2025vace}. Intermediate latent $\mathbf{z}_{t'}$ at timestamp $t'$ is generated as follow:
\begin{equation}
    \mathbf{z}_{t'} = (1-t')\mathbf{z}_0 + t' \mathbf{z}_1,
\end{equation}
where $\mathbf{z}_0$ denotes noise latent and $\mathbf{z}_1$ denotes ground truth target video in our setting.
Then, the flow matching loss is calculated as follows:
\begin{equation}
\mathcal{L}_{\mathrm{FM}} = \mathbb{E}_{\mathbf{z}_0, \mathbf{z}_1, t'} \left[ \left\| \mathcal{R}_\text{VDM}(\mathbf{z}_{t'}, t', \{\mathbf{f}_l\}_{l=0}^{L_\mathcal{R}},\, \mathcal{C} ) - (\mathbf{z}_1 - \mathbf{z}_0) \right\|_2^2 \right].
\end{equation}
This formulation learns a continuous transport field between noise state to image space.

\paragrapht{Implementation details.}
For training, we use input frames at a resolution of $480 \times 832$ and adopt the Kubric~\cite{greff2022kubric} dataset with the same sampling strategy as in \ours. The model is optimized using AdamW~\cite{loshchilov2017decoupled} with a learning rate of $1\mathrm{e}{-4}$, and we apply LoRA~\cite{hu2022lora} with rank $128$. Training is conducted on 4 NVIDIA H100 GPUs with a per-GPU batch size of $1$ and $4$ gradient accumulation steps, for a total of $50\text{K}$ steps.

\subsection{Dataset Details}
\label{supsubsec:dataset}
\paragrapht{Spring.} Spring dataset~\cite{mehl2023spring} is the high-quality synthetic video with a large proportion of dynamic objects. Since only a single monocular video $\mathcal{V}$ is available, we sample $N$ frames as context input $\mathcal{V}_c$ and use frames whose timestamps fall within the range spanned by $\mathcal{T}_c$. 

Specifically, we use the 37 scenes in training split. we randomly sample a start index from the video and select $T=23$ consecutive frames. Of these, every other frame beginning from the first is designated as a context input, yielding $|\mathcal{T}_c| = 12$ context frames with a temporal stride of 2. Photometric and geometric losses are applied across all 23 frames. The non-context frames — which are never provided as input — require the model to predict scene appearance at unobserved timestamps, as they fall temporally between context frames. This design explicitly encourages the model to develop temporal interpolation capabilities, learning to synthesize geometrically and photometrically consistent Gaussians at arbitrary intermediate time moments rather than simply memorizing seen views.

\paragrapht{Kubric.}
Following~\cite{koo2025mv}, we construct our training dataset using the Kubric rendering engine~\cite{greff2022kubric}, generating 1,144 synthetic scenes. Each scene consists of $M=4$ multi-view videos $\{ \mathcal{V}_m \}_{m=1}^{M}$, where $\mathcal{V}_m = \{ I_{t,m} \}_{t=1}^{T}$ is the video captured along the $m$-th camera trajectory. During training, a single trajectory $m'$ is chosen as the context input $\mathcal{V}_c = \mathcal{V}_{m'}$, and use frames from all $M$ trajectories as novel-view rendering targets. 

However, we find that restricting the context to a single viewpoint frequently forces the model to extrapolate into unobserved regions, leading to training instability. To address this, we augment the context set by incorporating one frame from each of the $M$ viewpoints at a randomly sampled timestamp $t'$:
\begin{equation}
    \mathcal{V}_c \leftarrow \mathcal{V}_{m'} \cup
    \{ I_{t',m} \}_{m=1}^{M}.
\end{equation}
By momentarily exposing the model to all viewpoints at a shared timestamp, this augmentation anchors multi-view geometry at that instant, stabilizes training, and enables multi-view consistency supervision that would otherwise be inaccessible from monocular video alone.

\paragrapht{RealEstate10K.}
We additionally use the RealEstate10K dataset~\cite{zhou2018stereo}, a multi-view static dataset, to enhance geometric consistency through multi-view supervision. We follow the same sampling strategy as Spring and set the target timestamp equal to the frame index, since RealEstate10K consists of static monocular videos.
\section{Evaluation Protocol}
\label{supsec:evalpro}
\subsection{Evaluation Dataset}
\paragrapht{DyCheck.}
Following NeoVerse~\cite{yang2026neoverse}, we evaluate on 5 scenes that provide multi-view videos with valid camera poses: \textit{apple}, \textit{block}, \textit{paper-windmill}, \textit{spin}, and \textit{teddy}. For context input, we use Camera 0, the causally captured front-facing camera, sampling 32 frames with a temporal stride of 2 starting from the first frame. Novel views are then rendered from a static camera (Camera 1) across 63 consecutive frames with stride 1, where all rendered timestamps are fully spanned by the temporal range of the context frames, ensuring evaluation remains strictly within the interpolation regime. Following Shape of Motion~\cite{wang2025shape}, all metrics are computed exclusively over co-visible regions to avoid penalizing the model for regions that are fundamentally unobservable from the given context input.

\paragrapht{Aria Digital Twin.}
Following 4DGT~\cite{xu20254dgt}, we use 4 monocular videos from Aria Digital Twin~\cite{pan2023aria}:
\begin{itemize}
    \renewcommand{\labelitemi}{--}
    \item \textit{Apartment\_release\_multiuser\_cook\_seq141\_M1292} \item \textit{Apartment\_release\_multiskeleton\_party\_seq114\_M1292} \item \textit{Apartment\_release\_meal\_skeleton\_seq135\_M1292} \item
    \textit{Apartment\_release\_work\_skeleton\_seq137\_M1292}
\end{itemize}
We sample 16 context frames with a temporal stride of 6 starting from the first frame, and evaluate on 15 target frames with a temporal stride of 6 starting from the fourth frame, such that all evaluation timestamps are interleaved between context frames and thus lie strictly within the interpolation regime.

\paragrapht{TUM-Dynamics.}
We use 9 monocular videos which is all train split in TUM-Dynamics dataset~\cite{sturm2012benchmark}:
\begin{itemize}
    \renewcommand{\labelitemi}{--}
    \item \textit{rgbd\_dataset\_freiburg2\_desk\_with\_person} \item \textit{rgbd\_dataset\_freiburg3\_sitting\_halfsphere} \item \textit{rgbd\_dataset\_freiburg3\_sitting\_rpy} \item
    \textit{rgbd\_dataset\_freiburg3\_sitting\_static} \item \textit{rgbd\_dataset\_freiburg3\_sitting\_xyz} \item \textit{rgbd\_dataset\_freiburg3\_walking\_halfsphere} \item \textit{rgbd\_dataset\_freiburg3\_walking\_rpy} \item
    \textit{rgbd\_dataset\_freiburg3\_walking\_static} \item \textit{rgbd\_dataset\_freiburg3\_walking\_xyz}
\end{itemize}
Similar to Aria Digital Twin, we sample 16 context frames with a temporal stride of 6 starting from the first frame, and evaluate on 15 target frames with a temporal stride of 6 starting from the fourth frame, such that all evaluation timestamps are interleaved between context frames.

\paragrapht{NVIDIA.}
For the NVIDIA dataset~\cite{yoon2020novel}, we evaluate on 7 scenes: \textit{Balloon1}, \textit{Balloon2}, \textit{Jumping}, \textit{Playground}, \textit{Skating}, \textit{Truck}, and \textit{Umbrella}. We sample context frames with a temporal stride of 2 starting from the first frame, and evaluate on the remaining interleaved frames.

\subsection{Test-Time Pose Alignment}
As discussed in \cite{ye2024no}, methods that do not take ground-truth camera poses as input lack absolute scale information, and thus may predict scenes at arbitrary scales even given identical inputs. This scale ambiguity makes it impossible to fairly evaluate rendering quality against ground-truth poses, as a geometrically accurate prediction may nonetheless yield poor metrics due to scale misalignment. To address this, we follow NoPoSplat~\cite{ye2024no} and perform test-time pose alignment before evaluation, using the aligned poses to measure rendering metrics on the rendered images.

\subsection{Baselines}
\paragrapht{4DGT, MoVieS:}
4DGT~\cite{xu20254dgt} and MoVieS~\cite{lin2025movies} is the feed-forward 4D Gaussian splatting network based on LRM architecture~\cite{zhang2024gs}. 4DGT requires RGB inputs and its corresponding camera poses and timestamps. For camera poses, all camera poses are transformed into the coordinate frame of the first image. The resulting translations are then normalized by subtracting their mean and dividing by the maximum distance from the mean, ensuring all translated camera centers lie within a unit sphere. This normalization improves numerical stability during inference. For timestamps, we normalize each frame index by the frame rate (FPS) of the corresponding dataset to obtain time values in seconds.

\paragrapht{NeoVerse:}
NeoVerse~\cite{yang2026neoverse} is a feed-forward 4D Gaussian Splatting network built on the VGGT~\cite{wang2025vggt} backbone that operates without camera pose inputs. Since NeoVerse consists of feed-forward 4D Gaussian and video diffusion network, we evaluate only feed-foward 4D Gaussian model which generates Gaussians from input view distribution without any generation priors. However, the absence of pose supervision introduces scale ambiguity, causing the reconstructed scene scale to misalign with the ground-truth camera poses and making it difficult to fairly assess rendering quality. To resolve this, we apply the same test-time camera pose optimization used in our method to NeoVerse as well, ensuring a consistent and fair comparison.
\section{Additional Experiments and Ablation}
\label{supsec:exp}
\subsection{Diffusion-based Rendering Results}
To evaluate the performance after diffusion-based refinement, we measure the quality of novel view synthesis results produced by our VDM-based enhancement module using VBench~\cite{huang2024vbench}. 
Specifically, we first obtain novel-view rendered videos and then feed them into the video diffusion module for refinement.
The VBench results of the refined outputs are reported in Tab.~\ref{suptab:vbench}.

Despite being trained only on the Kubric dataset, a substantially \emph{smaller corpus}, and using a diffusion model with far \emph{fewer parameters}, our method outperforms existing diffusion-based baselines on geometry-related metrics (e.g., Backward Consistent, Temporal Flickering, and Motion Smoothness), thanks to the global motion and consistent geometry provided by C4G rendering. While the limited training data and reduced model capacity yield slightly lower image quality, the C4G-rendered outputs serve as a coarse yet geometrically reliable anchor for novel view synthesis, enabling the diffusion module to produce videos with strong geometric consistency.


\begin{table}[t]
    \centering
    \caption{\textbf{VBench~\cite{huang2024vbench} results for novel view generation.} We randomly collect 100 unseen in-the-wild videos and each rendered different camera trajectories.}
    \resizebox{\linewidth}{!}{
    \begin{tabular}{l|c|cccccc}
        \toprule
        Method& Backbone & Sub. Consist. & Back. Consists. & Temp. Flick. & Motion Smooth. & Aesth. Quality & Image Quality \\
        \midrule \midrule
        TrajectoryCrafter~\cite{yu2025trajectorycrafter} & CogVideoX-Fun-5B~\cite{yang2024cogvideox}&0.8543&0.9015&0.9229&0.9559&0.4840&0.6216\\
        NeoVerse~\cite{yang2026neoverse} & Wan2.1-T2V-14B~\cite{wan2025wan} &0.8940&0.9159&0.9220&0.9566&0.5063&0.6401\\
        \midrule
        Ours & Wan2.1-VACE-1.3B~\cite{wan2025wan} &0.8880&0.9179&0.9595&0.9774&0.4760&0.4483\\
        \bottomrule
    \end{tabular}}
    \label{suptab:vbench}
\end{table}

\begin{table}[t]
    \centering
    \caption{\textbf{Ablation studies on timestamp normalization.}}
    \begin{tabular}{l|ccc|ccc}
        \toprule
        \multirow{2}{*}{Methods} & \multicolumn{3}{c|}{DyCheck~\cite{gao2022monocular}} & \multicolumn{3}{c}{TUM-Dynamics~\cite{sturm2012benchmark}} \\
         & PSNR $\uparrow$ & SSIM $\uparrow$ & LPIPS $\downarrow$ & PSNR $\uparrow$ & SSIM $\uparrow$ & LPIPS $\downarrow$ \\
        \midrule \midrule
        $t-t_{\text{tgt}}$ &15.58&0.384&0.387&18.15&0.549&0.367\\
        Min-max Norm. &\textbf{15.73}&\textbf{0.393}&\textbf{0.382}&18.40&0.560&0.355 \\
        Ours & 15.54& 0.383& 0.385&\textbf{18.94}&\textbf{0.575}&\textbf{0.341}\\
        
        \bottomrule
    \end{tabular}
    \label{suptab:ablation_timestamp}
    
\end{table}

\begin{table}[t]
    \centering
    \caption{\textbf{Ablation studies on dynamic loss function.}}
    \begin{tabular}{c|ccc|ccc}
        \toprule
        \multirow{2}{*}{$\lambda_\text{dynamic}$} & \multicolumn{3}{c|}{DyCheck~\cite{gao2022monocular}} & \multicolumn{3}{c}{TUM-Dynamics~\cite{sturm2012benchmark}} \\
         & PSNR $\uparrow$ & SSIM $\uparrow$ & LPIPS $\downarrow$ & PSNR $\uparrow$ & SSIM $\uparrow$ & LPIPS $\downarrow$ \\
        \midrule \midrule
        1& 15.46& 0.373& 0.373& 18.71& 0.578& 0.317\\
        \textbf{3}& \textbf{15.54}& 0.383& \textbf{0.385}& \textbf{18.94}& \textbf{0.575}& \textbf{0.341}\\
        5& 15.50& \textbf{0.386}& 0.394& \textbf{18.94}& 0.572& 0.353\\
        10& 15.27& 0.380& 0.405& 18.82& 0.564& 0.368\\

        \bottomrule
    \end{tabular}
    \label{suptab:dynamic_mask}
    
\end{table}
\subsection{Timestamp Ablation}
We conduct an ablation study on timestamp normalization strategies. Beyond our proposed method which scales all timestamps such that the target timestamp maps to 1, we compare against two alternatives: (1) shift normalization, which subtracts the target timestamp from all timestamps so that the target maps to 0, and (2) min-max normalization, which normalizes all timestamps using the minimum and maximum of the context timestamps.

As shown in Tab.~\ref{suptab:ablation_timestamp}, all three strategies yield comparable performance on DyCheck, but our method achieves superior results on the Bonn dataset. We attribute this to the fundamental difference in evaluation protocol between the two benchmarks. DyCheck evaluates novel-view synthesis at the same timestamp across different viewpoints, meaning the model only needs to correctly identify the target view to render accurately, making it relatively robust to the choice of normalization. In contrast, Bonn requires temporal interpolation between context frames, where the relative temporal relationships between context and target timestamps become critical. In this setting, our normalization — which fixes the target timestamp as a constant reference point and expresses all context timestamps relative to it — provides the most stable and consistent temporal conditioning, leading to the best interpolation performance.



\subsection{Dynamic Mask Loss.}
Since our model is initialized from pretrained weights trained exclusively on static scenes, it initially struggles to reconstruct dynamic regions. To address this, we up-weight the loss in dynamic regions relative to static ones. Given a dynamic mask $M_{\text{dyn}}$, the reweighted MSE loss of photometric loss is defined as:
\begin{equation}
    \mathcal{L}_{\text{MSE}}(t,i) = \left(\lambda_{\text{dyn}} M_{\text{dyn},t}(i) + (1 - M_{\text{dyn},t}(i))\right) \| I_t(i) - \hat{I}_t(i) \|,
\end{equation}
where $\lambda_{\text{dyn}} > 1$ amplifies the gradient signal in dynamic regions and $i$ denotes the pixel indices. The dynamic mask $M_{\text{dyn},t}$ is obtained from an off-the-shelf segmentation model given corresponding RGB images $I_t$. As shown in Tab.~\ref{suptab:dynamic_mask}, this dynamic-aware loss weighting significantly improves reconstruction quality in motion regions.


\section{Tracking Experiments}
\begin{table}[t]
    \centering
    \caption{\textbf{Quantitative results of point tracking}}
    \resizebox{\linewidth}{!}{
    \begin{tabular}{l|cccc|cccc}
        \toprule
        \multirow{2}{*}{Method}& \multicolumn{4}{c|}{ADT~\cite{pan2023aria}} & \multicolumn{4}{c}{DriveTrack~\cite{balasingam2024drivetrack}} \\
        & $\mathrm{\Delta t=2}$ & $\mathrm{\Delta t=4}$ & $\mathrm{\Delta t=8}$ & $\mathrm{\Delta t=16}$ & $\mathrm{\Delta t=2}$ & $\mathrm{\Delta t=4}$ & $\mathrm{\Delta t=8}$ & $\mathrm{\Delta t=16}$\\
        \midrule \midrule
        4DGT~\cite{xu20254dgt} &91.24&80.77&75.44&68.84 &40.85&40.17&22.90&32.30\\
        MoVieS~\cite{lin2025movies} &91.02&78.27&62.22& 64.36 &\phantom{*}7.50&\phantom{*}7.02&\phantom{*}2.70&\phantom{*}5.60\\
        NeoVerse~\cite{yang2026neoverse} &\textbf{96.06}&\textbf{97.83}&\textbf{94.67}&\textbf{82.76} &\underline{47.77}&\underline{47.81}&\underline{33.26}&\underline{41.95}\\
        \midrule
        Ours &\underline{93.98}&\underline{88.43}&\underline{76.06}&\underline{71.86}&\textbf{55.51}&\textbf{52.00}&\textbf{37.19}&\textbf{49.89}\\
        \bottomrule
    \end{tabular}}
    \label{suptab:tracking}
\end{table}
\label{supsec:tracking}
To validate the scene motion understanding of \ours\ over pixel-wise methods, we conduct point tracking evaluation against other feed-forward dynamic 3DGS frameworks on the ADT dataset~\cite{pan2023aria}. We convert movements of Gaussian from query points into a tracking field by propagating their center positions across timestamps. We evaluate our model by varying the input stride $\mathrm{\Delta t} \in \{2, 4, 8, 16\}$ and report the $<\delta_{16}$ metric, which measures the fraction of predicted tracks within 16 pixels of the ground truth.

As shown in Tab.~\ref{suptab:tracking}, we evaluate point tracking by measuring the temporal consistency of Gaussian trajectories on the ADT~\cite{pan2023aria} and DriveTrack~\cite{balasingam2024drivetrack} datasets with various input stride $\mathrm{\Delta t} \in \{2,4,8,16\}$. Despite the inherent disadvantage that our model is not a pixel-wise estimator — meaning the nearest Gaussian may not perfectly align with the query point, introducing positional noise at initialization — our model nonetheless shows competitive results over all per-pixel baselines. This demonstrates that our model captures more geometrically and temporally coherent scene motion, such that even trajectories initialized from imperfectly aligned Gaussians yield more accurate tracking than pixel-aligned predictions from competing methods. While our model underperforms NeoVerse on the ADT dataset, we attribute this gap to a favorable bias in NeoVerse's evaluation rather than superior motion understanding. Specifically, ADT contains a large proportion of static query points, and since NeoVerse relies on VGGT point maps, its nearest-neighbor matching tends to associate static points with one another, artificially inflating tracking scores. In contrast, our model produces temporally consistent trajectories without such bias, making the comparison on static-heavy benchmarks less indicative of true dynamic scene understanding.

\section{Additional Qualitative Results}
\label{supsec:addqual}
\subsection{Additional Attention Visualization on C4G.}
\begin{figure}
    \centering
    \includegraphics[width=0.94\linewidth]{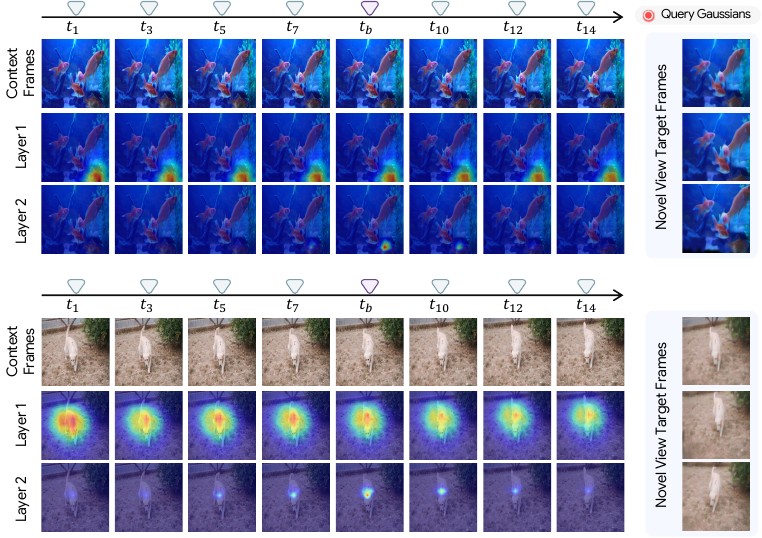}
    \caption{\textbf{Attention map visualization in dynamic regions.}}
    \label{fig:attention_vis_dynamic}
\end{figure}

We additionally provide the visualization results of attention map extended to Fig.~\ref{fig:analysis}. As shown in Fig.~\ref{fig:attention_vis_dynamic}, our model preserves its emergent attention properties~\cite{hong2022neural,hong2022integrative,han2025emergent} consistently across diverse input images, token positions, and timestamps. Examining the layer-wise attention patterns, we observe a clear functional specialization: Layer 1 attends broadly and consistently to spatially corresponding regions regardless of timestamp, while Layer 2 exhibits temporally-aware attention, where the attention intensity increases for tokens closer to the target timestamp and diminishes for those further away. This suggests a natural division of labor within the decoder, where Layer 1 primarily captures spatial correspondences~\cite{hong2021deep,cho2022cats++,cho2021cats,an2025cross} and Layer 2 progressively incorporates temporal context to refine the representation toward the target time moment.

\subsection{Analysis of Attention in VDM-based Rendering Enhancement Module}
\begin{figure}
    \centering
    \includegraphics[width=0.94\linewidth]{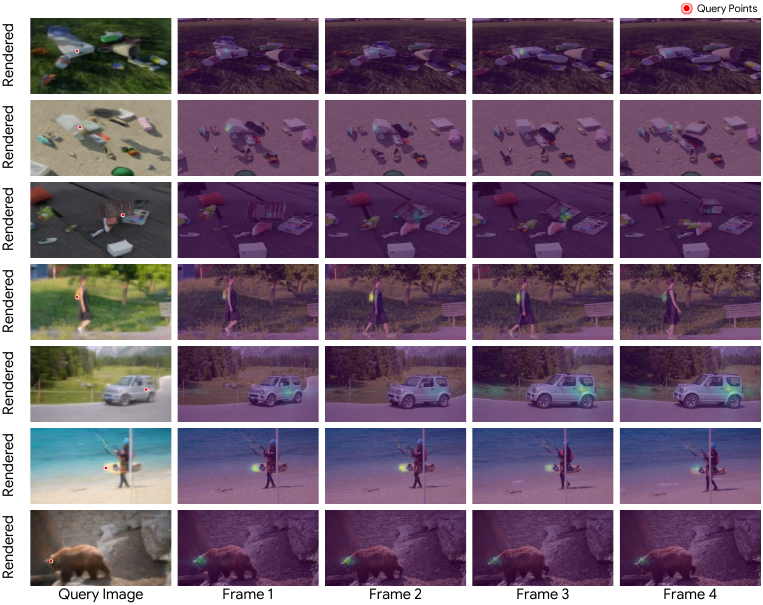}
    \caption{\textbf{Attention map visualization in VDM-based rendering enhancement module.}}
    \label{fig:attention_vis_vdm}
\end{figure}

We further investigate whether our VDM-based diffusion model leverages the reference images when enhancing the rendered images. To this end, we visualize the attention maps between the rendered images and the reference images within the control block in Fig.~\ref{fig:attention_vis_vdm}. As shown, the model attends to the reference images while refining the rendered ones, demonstrating that reference images play a crucial role in the enhancement process.

\clearpage
\section{Limitations}
\label{supsec:limitation}
Since our model reconstructs dynamic scenes solely from the input view distribution, it can only generate Gaussians for regions observed at least once across the context frames. Consequently, when rendering from extreme viewpoints far from the context cameras, or in regions fully occluded in all input views, the model produces holes with zero-valued pixels. While diffusion-based refinement can fill these regions, it tends to hallucinate plausible content rather than recover accurate real ground truth image. In addition, our model cannot extrapolate beyond the temporal range of the input views: since C4G is trained only on interpolated timestamps, it fails to predict accurate future Gaussians, which in turn prevents the diffusion module from synthesizing future frames, as it relies on the target views rendered by \ours.



\end{document}